\documentclass[12pt]{article}

\usepackage{graphicx}

\usepackage[utf8]{inputenc} 
\usepackage[T1]{fontenc}    
\usepackage[hidelinks, colorlinks=true,citecolor=blue]{hyperref} 
\usepackage{url}
\usepackage{booktabs}
\usepackage{amsfonts}
\usepackage{nicefrac}
\usepackage{microtype}
\usepackage[dvipsnames]{xcolor}
\usepackage[normalem]{ulem}
\usepackage{amsmath}
\usepackage{color}
\usepackage{natbib}
\usepackage{caption}
\usepackage{subcaption}
\usepackage{bm}
\usepackage[flushleft]{threeparttable}
\usepackage[shortlabels]{enumitem}
\usepackage{wrapfig}
\setlist{leftmargin=6mm}
\usepackage{multirow}
\usepackage{titlesec}
\usepackage{setspace}
\usepackage{pifont} 
\newcommand{\cmark}{\ding{51}} 
\newcommand{\xmark}{\ding{55}} 

\usepackage{algorithm}
\usepackage{algorithmic}
\newlength\myindent
\usepackage{amsthm}

\newtheorem{conjecture*}{Conjecture}
\theoremstyle{remark}

\allowdisplaybreaks

\newcommand{\calH}{\mathcal{H}}
\newcommand{\E}{\mathbb{E}}

\definecolor{olivegreen}{rgb}{0.1, 0.5, 0.0}

\newcommand{\rev}[1]{\textcolor{black}{#1}}
\newcommand{\revv}[1]{\textcolor{black}{#1}}

\newcommand{\blind}{1}

\begin{document}

\def\spacingset#1{\renewcommand{\baselinestretch}%
{#1}\small\normalsize} \spacingset{1}


\if 1\blind
{
  \title{\bf Deep Graph Kernel Point Processes over Networks}
  \author{Zheng Dong, Matthew Repasky\\
    H. Milton Stewart School of Industrial and Systems Engineering, \\ Georgia Institute of Technology\\
    and \\
    Xiuyuan Cheng\\
    Department of Mathematics, Duke University\\
    and \\
    Yao Xie \\
    H. Milton Stewart School of Industrial and Systems Engineering, \\ Georgia Institute of Technology}
  \maketitle
} \fi

\if0\blind
{
  \bigskip
  \bigskip
  \bigskip
  \begin{center}
    {\LARGE\bf Deep Graph Kernel Point Processes over Networks}
\end{center}
  \medskip
} \fi

\bigskip
\begin{abstract}
Point process models are widely used for continuous time discrete event data, where each data point includes time and additional information called “marks” --- locations, nodes, or event types. We present a new point process model for discrete event data over networks built upon Hawkes’ classic influence kernel-based formulation to capture the influence of historical events on future events’ occurrence. 
The key idea is to represent the influence kernel by Graph Neural Networks (GNN) to capture the underlying graph structure while leveraging the strong representation power of GNNs. Compared with prior works focusing on directly modeling the conditional intensity function using neural networks, our kernel presentation harnesses the repeated event influence patterns more effectively by combining statistical and deep models, achieving better model estimation/learning efficiency and superior predictive performance. Our work significantly extends the existing deep spatio-temporal kernel for point process data, which is inapplicable to our setting due to the fundamental difference in the nature of the observation space being Euclidean rather than a graph. We present comprehensive experiments on synthetic and real-world data to show the superior performance of the proposed approach against the state-of-the-art in predicting future events and uncovering the graph structure among data.
\end{abstract}


\noindent%
\vfill

\newpage
\spacingset{1.75} 
\section{Introduction}
\label{sec:introduction}

Discrete event data, in which each observation records an event time, location, and possibly additional information (``marks''), arise in a wide range of applications such as crime analysis \citep{zhu2022spatiotemporal}, healthcare \citep{wei2023granger}, earthquake modeling \citep{ogata1988statistical,zhu2021imitation}, and traffic systems \citep{zhu2021spatio}. In many modern applications, events are naturally associated with entities connected through an underlying (often latent) graph structure, motivating point process models defined over graphs, where nodes represent event types or locations and edges encode interdependencies among them \citep{perry2013point,moradi2020first,passino2023mutually}.

Classical self-exciting point process models introduced by Hawkes \citep{hawkes1971spectra} characterize event dependence through an influence kernel that governs how past events affect future occurrences. While traditional parametric kernels (e.g., exponential decay) offer interpretability, they often lack the flexibility needed to capture complex, non-stationary influence patterns. Recent advances in deep point process modeling address this limitation by parameterizing influence kernels using neural networks, leading to improved performance for temporal \citep{zhu2021deep,okawa2021dynamic}, spatio-temporal \citep{zhu2022neural}, and non-stationary kernels \citep{dong2022spatio}. A key advantage of kernel-based formulations is that they explicitly model repeated influence patterns, combining statistical structure with neural flexibility to achieve efficient estimation and strong predictive performance.

Despite these developments, most existing deep kernel point process models pay limited attention to the underlying graph structure in multi-dimensional settings, typically modeling only direct pairwise interactions through scalar parameters. \rev{Meanwhile, substantial progress has been made in graph signal processing and machine learning through graph filters and Graph Neural Networks (GNNs), which enable flexible modeling of multi-hop, nonlinear dependencies by explicitly incorporating graph topology \citep{bruna2013spectral,defferrard2016convolutional,gama2020graphs,leus2023graph}.} These advances suggest a natural opportunity to enrich point process kernels with graph-aware representations. However, integrating GNNs into point process models in a way that preserves statistical interpretability remains an open challenge.

In this paper, we propose a novel deep graph kernel framework for multi-dimensional point processes, in which the influence kernel is parameterized using GNNs to model complex event interactions over graphs. The proposed kernel is highly flexible, capable of capturing non-stationary, multi-hop, and inhibitory influences, as well as time-varying inter-node dynamics. Our framework accommodates a broad class of GNN architectures through localized graph filter bases and naturally extends to settings with unknown graph topology via attention mechanisms. By retaining an explicit kernel structure, the proposed approach preserves interpretability while substantially extending existing deep kernel models beyond Euclidean space to discrete graph domains.

Our main contributions are summarized as follows:
\begin{enumerate}
    \item We introduce a deep graph kernel framework for point process modeling that directly parameterizes the influence kernel using GNNs, enabling flexible modeling of non-stationary, multi-hop, and inhibitory interactions that are difficult to capture with traditional parametric kernels.
    \item The proposed framework supports a wide range of GNN architectures through localized graph filters and naturally handles both known and unknown graph structures, making it applicable to diverse domains such as traffic systems, healthcare, and social networks.
    \item Extensive experiments on synthetic and real-world datasets demonstrate that the proposed method improves latent dependency recovery and event prediction accuracy, consistently outperforming state-of-the-art baselines.
\end{enumerate}

\revv{A list of abbreviations used throughout the manuscript is provided in Appendix~H.}

\subsection{Related Work}

Deep learning approaches to point process modeling can be broadly divided into methods that directly model the conditional intensity function using neural networks and those that characterize event dependence through influence kernels. The former includes recurrent neural network (RNN)–based models~\citep{du2016recurrent,mei2017neural} and, more recently, self-attention–based approaches inspired by transformer architectures~\citep{zuo2020transformer,zhang2020self}. While these methods provide expressive models for the conditional intensity, they often lack the statistical interpretability associated with influence kernels and typically do not explicitly incorporate graph structure.

Modeling point processes with graph information has been explored in both classical and modern settings. Classical multivariate Hawkes processes~\citep{reinhart2018review} assume parametric kernels and represent node-to-node influence using scalar coefficients. Extensions such as the Group Network Hawkes Process~\citep{fang2023group} account for heterogeneous nodal characteristics through parametric formulations. Other approaches represent interactions as linear combinations of pre-specified parametric basis functions with low-rank constraints imposed on the coefficient matrix~\citep{tang2023multivariate}. In contrast, our approach adopts a functional perspective, representing the influence kernel itself using neural-network–parameterized basis functions with an explicit low-rank structure, enabling substantially richer interaction modeling.

Several recent works incorporate graph structure into deep point process models without explicitly using GNNs~\citep{yang2021transformer,zhang2021learning,zuo2020transformer,pan2023graphinformed}. For example, A-NHP~\citep{yang2021transformer} and THP-S~\citep{zuo2020transformer} employ attention-based mechanisms with additional constraints guided by the graph structure, which differ from Graph Attention Networks (GATs)~\citep{velickovic2017graph}. The Geometric Hawkes Process (GHP)~\citep{shang2019geometric} combines graph convolutional recurrent neural networks with Hawkes processes but relies on parametric, exponentially decaying kernels to model event influence. In contrast, our framework incorporates GNN architectures directly into the influence kernel without relying on parametric temporal bases, allowing flexible modeling of multi-hop, non-stationary dependencies over graphs.

Our work is also related to the extensive literature on graph signal processing and Graph Neural Networks (GNNs), which provide powerful tools for modeling structured data on graphs~\citep{bruna2013spectral,defferrard2016convolutional,gama2020graphs,leus2023graph}. GNNs have been applied to modeling temporal phenomena on graphs~\citep{longa2023graph,wu2020comprehensive}, using algebraic and spectral graph convolutions~\citep{bruna2013spectral,atwood2016diffusion,defferrard2016convolutional,kipf2016semi}, attention mechanisms~\citep{velickovic2017graph,brody2021attentive,dwivedi2020generalization}, and more recent architectures that combine local and global representations~\citep{cheng2020graph,rampavsek2022recipe,bianchi2021graph,zhu2021simple}. Outside the point process context, prior work has also addressed graph structure and influence learning via optimization of the graph Laplacian or adjacency matrix under various assumptions~\citep{xia2021graph,fatemi2021slaps,jin2020graph}.

Learning graph structure or influence in the context of point process modeling has also been studied~\citep{xu2016learning,zhou2013learninga,zhou2013learningb}. However, in these works, interactions between nodes are typically represented using learnable scalar parameters, which may be insufficient to capture complex, time-varying, and multi-hop dependencies in modern applications. Our model differs fundamentally by representing node-to-node influence as a learned function of time and graph structure through GNN-based kernels, enabling greater expressiveness while retaining interpretability.


\section{Background}

\paragraph{Self-exciting point processes.}
A self-exciting point process~\citep{reinhart2018review} models the occurrence of time-stamped discrete events whose intensity depends on the observed history. Let $\mathcal{H} = \{t_1, \dots, t_n\}$ be an event sequence with $t_i \in [0, T]$, and denote the history prior to time $t$ by $\mathcal{H}_t = \{t_i \mid t_i < t\}$. The conditional intensity function is defined as
\[
    \lambda(t) = \lim_{\Delta t \downarrow 0} \mathbb{E}\!\left[\mathbb{N}([t, t+\Delta t]) \mid \mathcal{H}_t\right] / \Delta t,
\]
where $\mathbb{N}$ denotes the counting measure. For notational simplicity, we suppress the explicit dependence of $\lambda(t)$ on $\mathcal{H}_t$.

The classical Hawkes process~\citep{hawkes1971spectra} models self-excitation additively via
\rev{
\[
    \lambda(t) = \mu + \sum_{t^\prime \in \mathcal{H}_t} k(t - t^\prime),
\]
}
where $\mu$ is the background intensity and $k:\mathbb{R}^+ \to \mathbb{R}^+$ is an influence kernel, often chosen to be exponential. We consider a more general kernel formulation~\citep{dong2022spatio},
\begin{equation}
    \lambda(t) = \mu + \sum_{t^\prime \in \mathcal{H}_t} k(t^\prime, t),
    \label{eq:intensity-with-kernel}
\end{equation}
where $k(t^\prime,t):\mathbb{R}^+\times\mathbb{R}^+\to\mathbb{R}$ allows for negative and non-stationary influences.

In a \emph{marked point process}, each event is associated with a mark $v\in V$, where $V$ is a discrete set of node indices. Let $\mathcal{H}_t=\{(t_i,v_i)\mid t_i<t\}$ denote the marked history. The conditional intensity then takes the form
\begin{equation}
    \lambda(t,v) = \mu + \sum_{(t^\prime,v^\prime)\in\mathcal{H}_t} k(t^\prime,t,v^\prime,v),
    \label{eq:conditional-intensity-with-kernel}
\end{equation}
where the influence kernel $k$ captures both temporal and nodal dependencies. A commonly used simplification expresses $k(t^\prime,t,v^\prime,v)$ as a product $a_{v,v^\prime} f(t-t^\prime)$, where $a_{v,v^\prime}$ encodes graph-based interactions and $f$ models temporal influence.

\paragraph{\rev{Graph convolution and Graph Neural Networks (GNNs).}}
Graph Neural Networks (GNNs)~\citep{wu2020comprehensive} process signals defined on graph nodes by iteratively applying graph convolutions and nonlinear transformations that incorporate graph structure. A graph convolutional layer computes a linear transformation of the input node features \revv{(i.e., graph signals)} followed by a node-wise nonlinearity.

Graph convolutions are commonly categorized as spectral or spatial. \rev{Given an adjacency matrix $A$ and degree matrix $D$, the graph Laplacian is $L := D-A$, with normalized form $\tilde L := I - D^{-1/2} A D^{-1/2}$.} Spectral graph convolutions apply a filter function $g_\theta$ to the spectrum of $L$, which can be expressed as $g_\theta(L)X$ using functional calculus. For example, ChebNet~\citep{defferrard2016convolutional} approximates $g_\theta$ using Chebyshev polynomial expansions.

When the polynomial degree is truncated, spectral methods admit an equivalent interpretation as spatial graph convolutions, in which information is propagated locally along edges. In spatial GNNs, the linear transformation at each layer aggregates features from a node’s local neighborhood, yielding localized graph filters that play a key role in capturing nodal dependencies.

We provide further details in Appendix~\ref{app:deep-graph-kernel-details} on representative GNN architectures compatible with the proposed deep graph kernel framework, including ChebNet~\citep{defferrard2016convolutional}, L3Net~\citep{cheng2020graph}, GAT~\citep{velickovic2017graph}, and the GPS Graph Transformer~\citep{rampavsek2022recipe}. In our experiments, we focus on L3Net and GAT, while the framework readily accommodates other GNN models.

\section{Point processes on graphs}

\subsection{Problem definition}

For a graph \( G = (V, E) \), let each node \( v \in V \) represent an event type. An undirected
edge between nodes \( u \) and \( v \) indicates the potential for interaction between type-\(u\)
and type-\(v\) events. These edges define the support of possible interactions but do not encode
directionality. In our learned model, the presence, absence, and direction of interactions along
these assumed edges are inferred from the data.

Our observations are a set of event sequences, $\mathcal S = \{\mathcal H^1, \ldots, \mathcal H^{M}\}$,
where each $\mathcal H^{s} = \{(t_i^s, v_i^s)\}_{i=1}^{n_s}$ is a collection of events $(t_i^s, v_i^s)$
occurring at node $v_i^s$ at time $t_i^s$, and \rev{$M$ is the number of trajectories.}
Our proposed graph point process aims to (i) jointly predict the times and types of forthcoming
events at any given time $t$ based on the observed historical data $\mathcal{H}_t$ and (ii) provide
an interpretable model of the event generation process by revealing the interdependence among
multiple event types and uncovering latent graph structure without prior information.
Toward this end, we adopt the statistical formulation of conditional intensity in
\eqref{eq:conditional-intensity-with-kernel} and introduce an influence kernel built on localized
graph filters in GNNs, aiming to characterize the relationship between any event pair
(\textit{e.g.}, excitation, inhibition through negative influence, or other network dynamic effects).

\subsection{Deep temporal graph kernel}\label{sec:deep_temporal_graph_kernel}

To go beyond simple parametric forms of the kernel while maintaining model efficiency, we use
a decomposition based on a generalization of Mercer's theorem for asymmetric kernels
\citep{mercer1909xvi, mollenhauer2020singular}.
Specifically, the influence kernel $k(t^\prime, t, v^\prime, v)$ in
\eqref{eq:conditional-intensity-with-kernel} is decomposed into basis kernel functions as follows:
\begin{equation}
    k(t', t, v', v) = \sum_{d=1}^D \sigma_d g_d(t', t-t^\prime) h_d(v', v), \quad t'<t,\ v,v' \in V,
    \label{eq:temporal-graph-decomposed-kernel}
\end{equation}
where $g_d: \mathbb R^+ \times \mathbb R^+ \rightarrow \mathbb R$, $h_d: V\times V\rightarrow \mathbb R$,
$d=1,\ldots, D$ are basis kernels in terms of event time and event type, respectively, and
$\sigma_d$ is the corresponding weight (``singular value'') for basis index $d$.
Instead of directly learning the full multi-dimensional event dependency, we simplify the task by
separately modeling different modes of temporal and graph-based dependence using distinct basis
functions, \rev{which will be represented by learnable neural networks.}
The weighted combination of basis kernels covers a broad range of influence kernels that can be
non-stationary and need not take the product form of temporal influence and nodal interactions.
While functional SVD is usually infinite-dimensional, in practice we truncate the decomposition and
use a low-rank approximation when the singular values $\sigma_d$ decay sufficiently fast (see an
example in Appendix~\ref{app:hyperparameter}).

The temporal basis kernels are designed to capture heterogeneous temporal dependencies between
past and future events. The parameterization of temporal kernels $\{g_d\}_{d=1}^{D}$ using
displacements $t-t^\prime$ (rather than $t$) provides a low-rank way to approximate general kernels
\citep{dong2022spatio}. We approximate $\{g_d\}_{d=1}^{D}$ using shared basis functions:
\[
    g_d(t^\prime, t-t^\prime) = \sum_{l=1}^{L}\beta_{dl}\psi_l(t^\prime)\varphi_l(t-t^\prime),
    \quad \forall d = 1, \dots, D.
\]
Here $\{\psi_l, \varphi_l: [0, T] \to \mathbb{R}\}_{l=1}^{L}$ are two sets of basis functions that
characterize the temporal impact of an event occurring at $t^\prime$ and the spread pattern over
$t-t^\prime$, respectively, and $\beta_{d,l}$ is the corresponding weight.
The basis functions $\{\psi_l, \varphi_l\}_{l=1}^{L}$ are represented by independent fully-connected
neural networks \rev{with the same architecture and different parameters estimated from data.
This representation leverages neural approximation capabilities and moves beyond restrictive
parametric basis choices.}

\subsection{Graph basis kernel with localized graph filters}
\label{sec:graph-basis-kernel}

We further represent graph basis kernels using localized graph filters that support graph convolution.
Specifically, the basis kernels $\{h_d\}_{d=1}^{D}$ are represented as
\[
    h_d(v^\prime, v) = \sum_{r=1}^{R}\gamma_{dr}B_r(v^\prime, v), \quad \forall d = 1, \dots, D,
\]
where $\{B_r(v^\prime, v): V \times V \to \mathbb{R}\}_{r=1}^{R}$ are $R$ bases of localized graph
filters, and $\gamma_{dr}$ is the corresponding weight for each $B_r$.
The bases can be constructed using either algebraic or spectral graph convolution approaches.

Combining the temporal and graph expansions above, the influence kernel \( k \) can be represented as
\begin{equation}
    k(t^\prime, t, v^\prime, v) = \sum_{r=1}^{R}\sum_{l=1}^{L}\alpha_{rl}\psi_l(t^\prime)\varphi_l(t - t^\prime)B_r(v^\prime, v),
    \label{eq:kernel-formulation}
\end{equation}
where \rev{\( \alpha_{rl} = \sum_{d=1}^{D}\sigma_d\beta_{dl}\gamma_{dr} \)}.
\rev{In model fitting, the parameters \( \alpha_{rl} \), along with the basis functions \( \psi_l \),
\( \varphi_l \), and \( B_r \), are all learned from data. Note that we allow the kernel
\( k(t^\prime, t, v^\prime, v) \) to take negative values. To ensure the non-negativity of the
conditional intensity function, we include a log-barrier term in the learning objective; see
Section~\ref{sec:model-estimation} for details.}
To demonstrate the flexibility of our framework, we present two example GNN architectures in the
numerical experiments (Section~\ref{sec:experiments}): \textit{L3Net}~\citep{cheng2020graph}, which
unifies algebraic and spectral graph convolutions, and Graph Attention Networks \textit{GAT}~\citep{velickovic2017graph}, which can learn the graph topology directly and is suitable when the graph
structure is unknown. Our framework is compatible with a wide range of GNN layers, including those
in \textit{ChebNet}~\citep{defferrard2016convolutional} and the \textit{GPS Graph Transformer}~\citep{dwivedi2020generalization}. An example of incorporating L3Net is provided in
Appendix~\ref{app:L3Net}, and additional technical details on other GNN architectures are given in
Appendix~\ref{app:deep-graph-kernel-details}.

\paragraph{Choice of basis functions.}
\revv{The choice of basis functions depends on the application setting.
For the temporal kernel, we use multilayer perceptrons to parameterize
$\{\psi_l, \phi_l\}$ due to their flexibility in modeling nonstationary temporal effects.
For the graph kernel, when the graph topology is known, we use localized graph filters
such as L3Net; when the graph structure is unknown, we adopt GAT
to learn latent graph connectivity from data. Other GNN-based graph filters can be incorporated
within the same framework (Appendix~\ref{app:deep-graph-kernel-details}).}

\revv{Although the kernel decomposition in \eqref{eq:temporal-graph-decomposed-kernel} involves a nominal rank parameter $D$,
this parameter does not appear explicitly in the final kernel representation in
\eqref{eq:kernel-formulation}. The singular values and expansion coefficients are absorbed
into the learned coefficients $\{\alpha_{rl}\}$, and the model is fully specified by the
temporal rank $L$ and graph rank $R$. As a result, only $L$ and $R$ need to be selected in practice.}

We remark on two key advantages of the proposed influence kernel \( k \).
First, it enables the use of a wide range of spectral and algebraic GNN filter bases, allowing the
model to capture both local and multi-hop inter-node influence patterns.
Second, the framework reduces the number of trainable parameters to \( \mathcal{O}(RC|V|) \),
where \( C \) denotes the average neighborhood size of graph nodes.
\revv{Here, the complexity refers to the number of trainable parameters in the graph kernel representation; empirical training times are reported in Appendix~\ref{app:efficient-computation}.}

\paragraph{Choice of rank.}
\revv{In practice, the kernel ranks \( L \) and \( R \) are treated as model hyperparameters
and selected using validation log-likelihood, with additional guidance from the magnitude
of the learned coefficients \( \{\alpha_{rl}\} \).}
Detailed empirical guidance and examples are provided in Appendix~\ref{app:hyperparameter}.

\subsection{Model estimation}
\label{sec:model-estimation}

Previous studies on point processes primarily employ two types of loss functions for model estimation:
the (negative) log-likelihood (NLL)~\citep{dong2023non, reinhart2018review} and the least squares (LS)
loss~\citep{bacry2020sparse, cai2022latent}. In our framework, we consider both loss functions and provide
detailed derivations in Appendix~\ref{app:loss-derivation}. For the observed data, \rev{we use
\( \lambda^{(s)}(t, v) \) to denote the conditional intensity function for the \( s \)-th sequence, which
depends on the same set of model parameters
\( \theta = \{ \alpha_{rl}, \theta^\psi_l, \theta^\phi_l, \theta^B_r : r=1,\dots,R,\, l=1,\dots,L \} \),
but varies due to differences in event history.}

\paragraph{Negative Log-Likelihood (NLL).}
The model parameters can be estimated by minimizing the negative log-likelihood~\citep{reinhart2018review}:
\[
    \ell_{\text{NLL}}(\theta):= \frac{1}{M}\sum_{s=1}^{M} \left(
    \sum_{v \in V}\int_0^T\lambda^{(s)}(t, v)\,dt
    -
    \sum_{i=1}^{n_s} \log \lambda^{(s)}\left(t_i^s, v_i^s\right)
    \right ).
\]
Note that the model parameters $\theta$ are incorporated through $\lambda^{(s)}$.

\paragraph{Least Square (LS) loss.}
Another commonly used loss is the least squares (LS) loss, which is based on matching the conditional
intensity function to its expected value:
\[
    \ell_{\text{LS}}(\theta)
    := \frac{1}{M} \sum_{s=1}^{M} \left( \sum_{v \in V} \int_{0}^{T} \lambda^{(s)}(t, v)^2 \, dt
    - \sum_{i=1}^{n_s} 2\lambda^{(s)}(t_i^s, v_i^s) \right).
\]
The least squares loss can be derived from the principle of empirical risk minimization~\citep{geer2000empirical}.
\rev{Note that the MLE and LS losses differ in that MLE requires evaluation of the logarithm of the conditional
intensity function, which can cause numerical issues when the intensity is close to zero.}

Since negative values of the influence kernel are allowed for learning inhibiting effects, additional processing
is required to ensure the non-negativity of the conditional intensity function. To achieve this, we leverage the
log-barrier method for optimization in point processes~\citep{dong2022spatio}. The learning objective function is
formulated as
\[
    \begin{aligned}
        &\min_{\theta} \mathcal{L}_{\text{NLL}}(\theta) := \ell_{\text{NLL}}(\theta) + \frac{1}{w}p(\theta, b),
        \quad \text{and} \quad
        \min_{\theta} \mathcal{L}_{\text{LS}}(\theta) := \ell_{\text{LS}}(\theta) + \frac{1}{w}p(\theta, b),
    \end{aligned}
\]
for the two loss functions, respectively. Here
\[
p(\theta, b) = - \frac{1}{M|\mathcal{U}_{\text{bar}, t} \times V|}
\sum_{s=1}^{M}\sum_{t \in \mathcal{U}_{\text{bar}, t}}\sum_{v \in V}\log(\lambda(t, v) - b)
\]
is the log-barrier function with regularization parameter $w >0$.
The set $\mathcal{U}_{\text{bar}, t}$ is a dense grid over $[0, T]$ on which the intensity function is penalized,
and $b$ is a lower bound of the intensity value.
\rev{In the experiments, the value of \( b \) is chosen based on empirical evaluation of the conditional intensity
function. It is initially set to a negative value during the early stages of training to ensure the feasibility of
the logarithm in the loss function, and is gradually tuned toward zero as learning progresses.}
Both loss functions can be efficiently computed numerically, as illustrated in Appendix~\ref{app:efficient-computation}.

\revv{Cross-validation is performed at the trajectory or subsequence level (rather than at the event level)
to account for temporal dependence; see Appendix~\ref{app:hyperparameter} for details.}

\section{Experiment}\label{sec:experiments}

In this section, we compare our method, referred to as \texttt{GraDK}, with seven state-of-the-art point process methods on both synthetic and real data sets.
\revv{To improve readability, we highlight representative results in the main text and defer additional experiments, ablation studies, and robustness checks to the appendix.}

\paragraph{Baselines.}
We include two groups of baselines with distinctive model characteristics. Models in the first group treat the associated node information of each event as one-dimensional event marks without considering the graph structure, including (i) Recurrent Marked Temporal Point Process (\texttt{RMTPP})~\citep{du2016recurrent} that uses a recurrent neural network to encode dependence through time; (ii) Fully Neural Network model (\texttt{FullyNN})~\citep{omi2019fully} that models the cumulative distribution via a neural network; and (iii) Deep Non-Stationary Kernel~\citep{dong2022spatio} with a low-rank neural marked temporal kernel (\texttt{DNSK-mtpp}). The second group includes two models that encode the latent graph structure information when modeling event times and marks, including (iv) Structured Transformer Hawkes Process (\texttt{THP-S})~\citep{zuo2020transformer} and (v) Graph Self-Attentive Hawkes Process (\texttt{SAHP-G})~\citep{zhang2020self} with a given graph structure, which both use self-attention mechanisms to represent the conditional intensity. More comparisons are presented in Appendix~\ref{app:comp-with-baselines}.

\paragraph{Experimental setup.}
We demonstrate the versatility of the proposed method to various GNN architectures and loss functions by considering: (i) \texttt{GraDK} with L3Net and NLL (\texttt{GraDK+L3net+NLL}), (ii) \texttt{GraDK} with L3Net and LS (\texttt{GraDK+L3net+LS}), and (iii) \texttt{GraDK} with GAT and NLL (\texttt{GraDK+GAT+NLL}). Details about the experimental setup and model architectures are provided in Appendix~\ref{app:additional-exp}.

\begin{table*}[!t]
  \caption{Synthetic data results.}
  \centering
  \begin{threeparttable}
    \begin{tabular}{ccccccc}
    \toprule
    \toprule
    \multirow{2}{*}{} & \multicolumn{3}{c}{\multirow{2}{*}{\shortstack{16-node graph \\ with \textit{2-hop} influence}}} & \multicolumn{3}{c}{\multirow{2}{*}{\shortstack{50-node graph \\ with \textit{non-separable} kernel}}} \\ \\
    \cmidrule(lr){2-4} \cmidrule(lr){5-7}
    \multirow{2}{*}{Model} & \multirow{2}{*}{\shortstack{Testing \\ $\ell$}} & \multirow{2}{*}{\shortstack{Time \\ MAE}} & \multirow{2}{*}{\shortstack{Type \\ KLD}} & \multirow{2}{*}{\shortstack{Testing \\ $\ell$}} & \multirow{2}{*}{\shortstack{Time \\ MAE}} & \multirow{2}{*}{\shortstack{Type \\ KLD}} \\ \\
    \midrule
    \texttt{RMTPP} & $-7.239_{(0.193)}$ & $0.301$ & $0.142$ & $-9.560_{(0.513)}$ & $41.606$ & $0.371$ \\
    \texttt{FullyNN} & $-3.347_{(0.018)}$ & $0.198$ & $0.018$ & $-3.098_{(0.176)}$ & $25.080$ & $0.054$ \\
    \texttt{DNSK-mtpp} & $-3.005_{(0.002)}$ & $0.085$ & $0.002$ & $-1.595_{(0.003)}$ & $1.043$ & $0.005$ \\
    \midrule
    \texttt{THP-S} & $-3.079_{(0.004)}$ & $0.108$ & $0.011$ & $-1.587_{(0.004)}$ & $3.110$ & $0.003$ \\
    \texttt{SAHP-G} & $-3.036_{(0.008)}$ & $0.155$ & $0.005$ & $-1.591_{(0.005)}$ & $3.482$ & $0.009$ \\
    \midrule
    \texttt{GraDK+L3net+NLL} & $-2.995_{(0.004)}$ & $0.028$ & $0.002$ & $-1.583_{(0.002)}$ & $\textbf{0.244}$ & $\textbf{<0.001}$ \\
    \texttt{GraDK+L3net+LS} & $\textbf{--2.993}_{(0.003)}$ & $\textbf{0.001}$ & $\textbf{0.001}$ & $\textbf{--1.582}_{(0.002)}$ & $0.253$ & $\textbf{<0.001}$ \\
    \texttt{GraDK+GAT+NLL} & $-2.997_{(0.005)}$ & $0.148$ & $\textbf{<0.001}$ & $-1.584_{(0.003)}$ & $0.354$ & $0.003$ \\
    \bottomrule
    \bottomrule
  \end{tabular}
  \begin{tablenotes}
  \item *Numbers in parentheses are standard errors for three independent runs.
  \end{tablenotes}
  \end{threeparttable}
  \label{tab:synthetic-data-results}
\end{table*}

\subsection{Synthetic data}
\label{sec:synthetic-data}

We consider 5 synthetic data sets: (i) a non-stationary temporal kernel on a 3-node graph with negative influence; (ii) a non-stationary temporal kernel on a 16-node graph with 2-hop graph influence; (iii) an exponentially decaying temporal kernel on a 50-node graph with central nodes; (iv) a non-separable diffusion kernel on a 50-node graph with central nodes; and (v) an exponentially decaying temporal kernel on a 225-node graph.
\revv{Each synthetic dataset consists of 1{,}000 independent event sequences, with time window length 50 for datasets (i)–(ii) and 10 for datasets (iii)–(v).}
Additional simulation details (including average sequence lengths) are provided in Appendix~\ref{app:additional-exp}.

\subsubsection{Known graph topology}

We start by showing the performance of \texttt{GraDK}, where the graph topology is known.

\begin{figure}[!t]
    \centering
    \begin{subfigure}[h]{.99\linewidth}
    {\includegraphics[width=\linewidth]{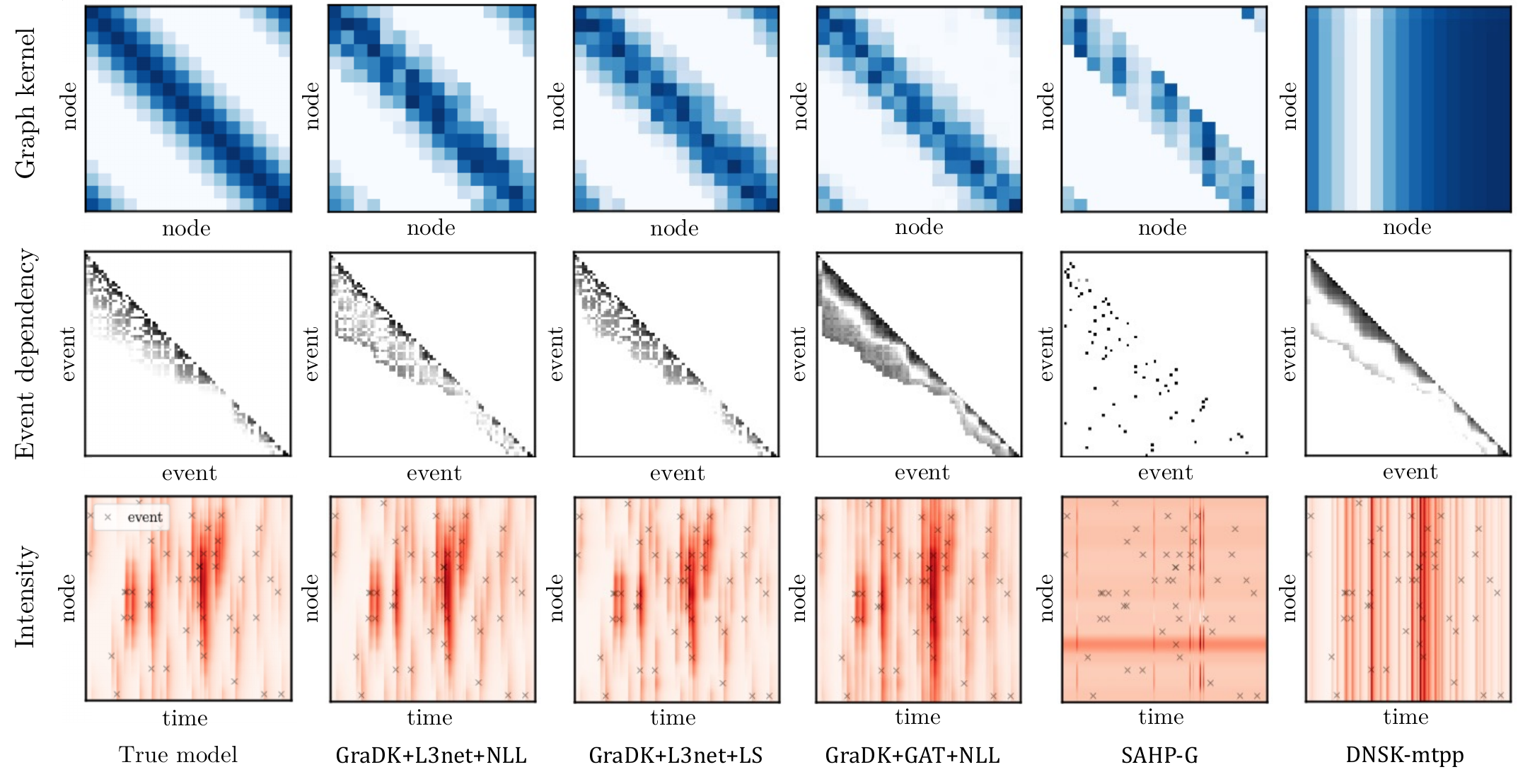}}
    \end{subfigure}
    \vspace{-0.05in}
    \caption{Graph kernel, inter-event dependence, and conditional intensity recovery for the 16-node synthetic data set with 2-hop graph influence. The first column reflects the ground truth, while the subsequent columns reflect the results obtained by \texttt{GraDK} (with three architectures), \texttt{SAHP-G}, and \texttt{DNSK}, respectively.}
    \label{fig:16-node-synthetic-data-vis}
\end{figure}

\paragraph{Kernel and intensity recovery.}
Figure~\ref{fig:16-node-synthetic-data-vis} shows the recovered graph kernel by each method for the synthetic data on a 16-node ring graph with 2-hop influence. Our method and \texttt{DNSK} directly learn the kernel, and the graph kernel in \texttt{SAHP-G} is constructed as in the original paper~\citep{zhang2020self} by computing the empirical mean of the learned attention weights between nodes. The proposed method recovers the self-exciting and multi-hop influence patterns in the kernel, while \texttt{SAHP-G} largely captures dependencies within one-hop neighbors. The corresponding recovered conditional intensities are shown in the bottom row of Figure~\ref{fig:16-node-synthetic-data-vis}.

\begin{figure}[!t]
    \centering
    \begin{subfigure}[h]{.7\linewidth}
    \includegraphics[width=1\linewidth]{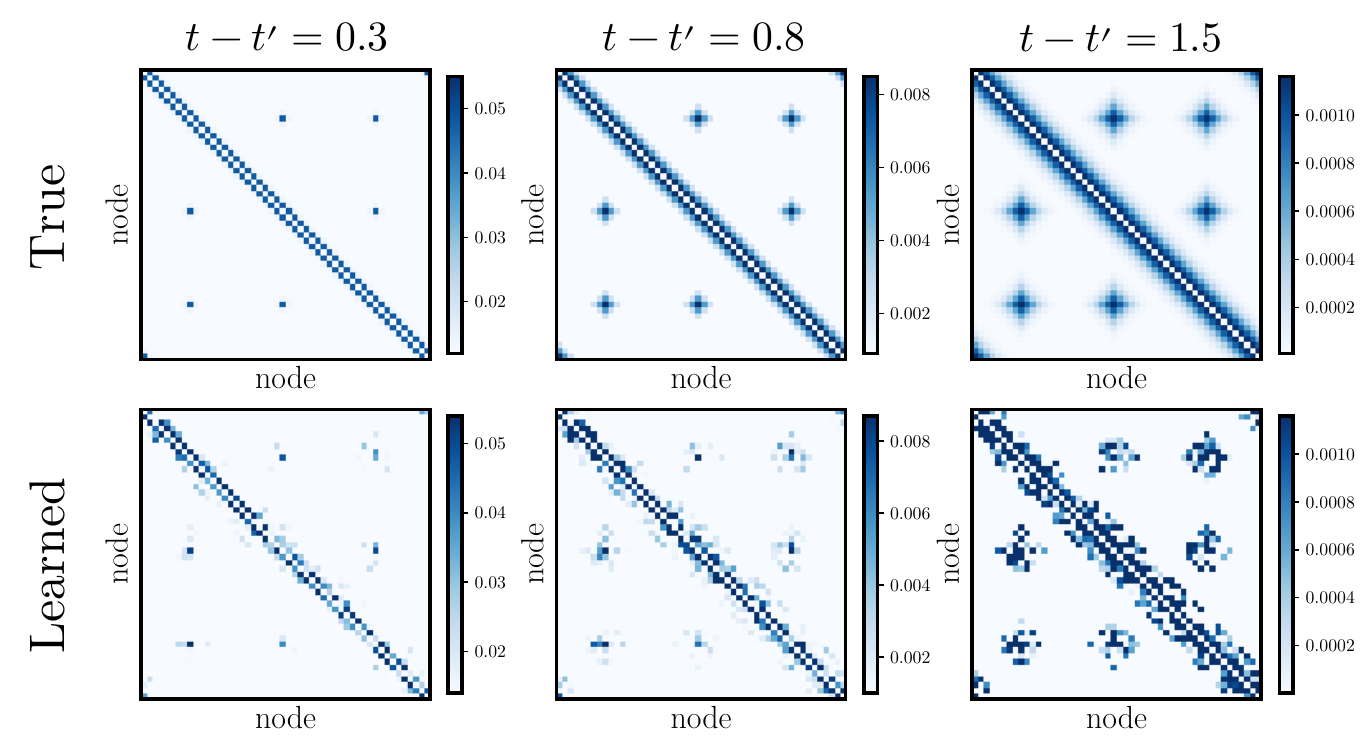}
    \end{subfigure}
\caption{Dynamic influence among nodes learned by \texttt{GraDK} on synthetic data generated by the 50-node graph with \textit{non-separable diffusion kernel}. Both true and learned kernels are evaluated at three time lags of $0.3, 0.8$, and $1.5$.}
\label{fig:dynamic-kernel}
\end{figure}

\paragraph{Event dependency.}
The second row of Figure~\ref{fig:16-node-synthetic-data-vis} shows event dependencies characterized by the influence kernel \eqref{eq:intensity-with-kernel} for \texttt{GraDK}, \texttt{DNSK}, and the true model. For \texttt{SAHP-G}, event dependency is indicated by the scaled self-attention weight (Equation 10 in~\citep{zhang2020self}). We observe that \texttt{SAHP-G} can discover long-term dependencies, while \texttt{DNSK} captures short-term decay but may miss longer-term effects for certain event types; the proposed method captures both effects. Additional synthetic results are reported in Appendix~\ref{app:additional-exp}.

\paragraph{Predictive performance.}
Apart from assessing the fitted log-likelihood ($\ell$) of the testing data, we also evaluate predictive performance. For each data set, we generate $M=100$ event sequences using each learned model (one of three independent runs) and evaluate two predictive metrics~\citep{juditsky2020convex}: (i) the mean absolute error of predicted event frequency (\textit{Time MAE}) compared to that in the testing data, and (ii) the Kullback--Leibler Divergence of predicted event types (\textit{Type KLD}). Results in Table~\ref{tab:synthetic-data-results} (and Table~\ref{tab:app-synthetic-data-results}) show strong predictive performance of \texttt{GraDK} across datasets.

\paragraph{Dynamic graph influence.}
Figure~\ref{fig:dynamic-kernel} illustrates that the proposed method can recover time-varying interactions among nodes for the non-separable diffusion kernel. The quantitative results in Table~\ref{tab:synthetic-data-results} further support the effectiveness of \texttt{GraDK} in capturing dynamic influence patterns.

\paragraph{Comparison of NLL and LS.}
Both NLL and LS losses are effective for learning the graph kernel, with LS often exhibiting slightly better numerical stability in our experiments. As shown in Appendix~\ref{app:efficient-computation}, both losses can be computed efficiently with similar computational complexity \( \mathcal{O}(n) \), where \( n \) is the total number of events, and their runtimes are comparable.

\subsubsection{Unknown graph topology}

We now demonstrate recovery of the event dependency structure when the graph topology is unknown, using GAT in the graph kernel. In the \texttt{GraDK+GAT+NLL} experiments, we initialize the model with a fully connected graph, and learned attention weights induce directional influences between nodes. As shown in Figure~\ref{fig:16-node-synthetic-data-vis}, the recovered graph kernels and event dependencies produced by \texttt{GraDK+GAT+NLL} closely match the ground truth.

\subsection{Real data}
\label{sec:real_data_experiment}

We test the performance of our proposed approach on real-world point process data. Since applications of graph point processes involve discrete events over networks with asynchronous time information, most traditional benchmark graph datasets are not directly applicable.
\rev{We consider four datasets: (i) Atlanta traffic congestion events from 5 sensors (24-hour sequences), (ii) Sepsis-associated derangements and sepsis onset events (13 event types, 24-hour per patient), (iii) California wildfire events clustered into 25 locations (one-year sequences), and (iv) Valencia theft events near 52 pubs (quarter-year sequences).}
Further dataset details are provided in Appendix~\ref{app:additional-exp}.

\begin{figure}[!t]
    \centering
    \begin{subfigure}[h]{.4\linewidth}
    \includegraphics[width=1\linewidth]{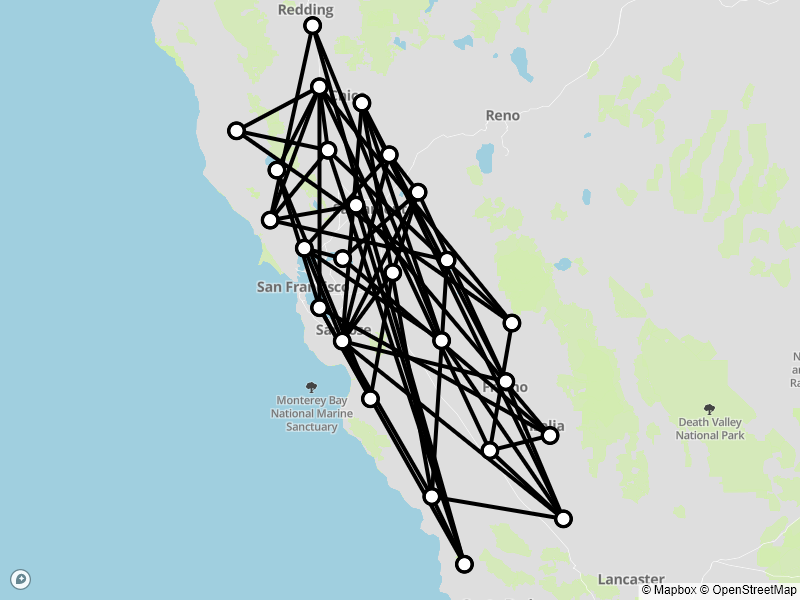}
    \caption{Wildfire graph}
    \end{subfigure}
    \hspace{0.02in}
    \begin{subfigure}[h]{.4\linewidth}
    \includegraphics[width=1\linewidth]{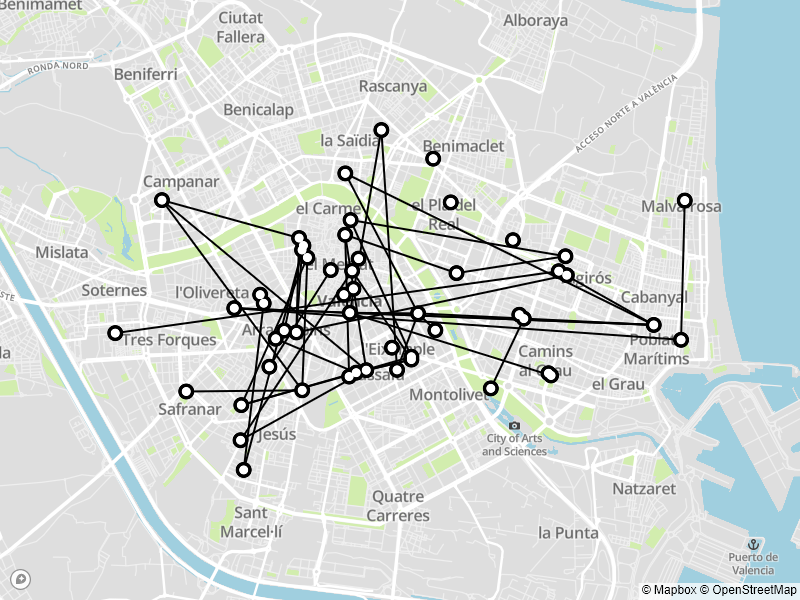}
    \caption{Theft graph}
    \end{subfigure}
\caption{Graph structures for California wildfire data and the Valencia sustraccion (theft) data.}
\label{fig:real-data-graphs}
\end{figure}

\subsubsection{Ablation study}
\label{sec:ablation-study}

For real-world applications involving discrete event data observed in geographic space, spatio-temporal point processes (STPPs) with Euclidean-distance-based influence kernels are commonly used. To evaluate the effectiveness of the proposed method, we compare \texttt{GraDK} with three STPP baselines on wildfire and theft datasets originally observed in Euclidean space. The three baselines are: (i) the Multivariate Hawkes Process (\texttt{MHP})~\citep{hawkes1971spectra}, (ii) the Epidemic-Type Aftershock Sequence (\texttt{ETAS}) model~\citep{ogata1988statistical}, and (iii) the Deep Non-Stationary Kernel model with a low-rank spatio-temporal kernel (\texttt{DNSK-stpp}).
\revv{Including \texttt{MHP} provides a classical parametric Hawkes benchmark when parametric assumptions are appropriate.}

\begin{table}[!t]
  \caption{Ablation study: comparison with Spatio-Temporal Point Processes (STPPs); the graph structure is shown in Figure~\ref{fig:real-data-graphs}.}
  \centering
  \begin{threeparttable}
  \begin{tabular}{cccccccc}
    \toprule
    \toprule
    & & \multicolumn{3}{c}{\multirow{2}{*}{\shortstack{Wildfire\\(25 nodes)}}} & \multicolumn{3}{c}{\multirow{2}{*}{\shortstack{Theft\\(52 nodes)}}} \\ \\
    \cmidrule(lr){3-5} \cmidrule(lr){6-8}
    \multirow{2}{*}{Model} & \multirow{2}{*}{\shortstack{\# Params.}} & \multirow{2}{*}{\shortstack{Testing \\ $\ell$}} & \multirow{2}{*}{\shortstack{Time \\ MAE}} & \multirow{2}{*}{\shortstack{Type \\ KLD}} & \multirow{2}{*}{\shortstack{Testing \\ $\ell$}} & \multirow{2}{*}{\shortstack{Time \\ MAE}} & \multirow{2}{*}{\shortstack{Type \\ KLD}} \\ \\
    \midrule
     \texttt{MHP} & $625$ & $-3.846_{(0.003)}$ & $1.103$ & $0.072$ & $-3.229_{(0.005)}$ & $1.912$ & $0.483$ \\
    \texttt{ETAS} & $2$ & $-3.702_{(0.002)}$ & $1.134$ & $0.385$ & $-3.049_{(0.001)}$ & $3.750$ & $0.685$\\
    \texttt{DNSK-stpp} & $4742$ & $\textbf{--3.647}_{(0.005)}$ & $0.861$ & $0.214$ & $-3.004_{(0.002)}$ & $1.342$ & $0.600$\\
    \texttt{GraDK} & $411$ & $-3.650_{(0.002)}$ & $\textbf{0.580}$ & $\textbf{0.018}$ & $\textbf{--2.998}_{(0.002)}$ & $\textbf{0.127}$ & $\textbf{0.292}$\\
    \bottomrule
    \bottomrule
  \end{tabular}
  \begin{tablenotes}
  \item *Numbers in parentheses are standard errors for three independent runs.
  \end{tablenotes}
  \end{threeparttable}
  \label{tab:ablation-study}
\end{table}

Table~\ref{tab:ablation-study} shows that \texttt{GraDK} performs competitively in fit and yields strong predictive performance on both datasets.

\subsubsection{Comparison with baselines}

\begin{table*}[!t]
  \caption{Real data results.}
  \vspace{-0.1in}
  \centering
  \begin{threeparttable}
  \begin{tabular}{ccccccc}
    \toprule
    \toprule
    & \multicolumn{3}{c}{\multirow{2}{*}{\shortstack{Sepsis\\(13 nodes)}}} & \multicolumn{3}{c}{\multirow{2}{*}{\shortstack{Theft\\(52 nodes)}}} \\ \\
    \cmidrule(lr){2-4} \cmidrule(lr){5-7}
    \multirow{2}{*}{Model} & \multirow{2}{*}{\shortstack{Testing \\ $\ell$}} & \multirow{2}{*}{\shortstack{Time \\ MAE}} & \multirow{2}{*}{\shortstack{Type \\ KLD}} & \multirow{2}{*}{\shortstack{Testing \\ $\ell$}} & \multirow{2}{*}{\shortstack{Time \\ MAE}} & \multirow{2}{*}{\shortstack{Type \\ KLD}} \\ \\
    \midrule
    \texttt{RMTPP} & $-7.395_{(0.613)}$ & $3.159$ & $1.124$ & $-11.496_{(1.474)}$ & $5.871$ & $0.124$ \\
    \texttt{FullyNN} & $-4.666_{(0.133)}$ & $0.302$ & $0.345$ & $-3.468_{(0.068)}$ & $6.457$ & $1.169$ \\
    \texttt{DNSK-mtpp} & $-4.088_{(0.012)}$ & $0.712$ & $0.057$ & $-3.347_{(0.012)}$ & $0.507$ & $0.177$ \\
    \midrule
    \texttt{THP-S} & $-3.543_{(0.008)}$ & $0.113$ & $0.067$ & $-2.982_{(<0.001)}$ & $0.739$ & $0.189$ \\
    \texttt{SAHP-G} & $\textbf{--3.134}_{(0.106)}$ & $0.858$ & $1.061$ & $\textbf{--2.970}_{(0.032)}$ & $0.464$ & $0.096$ \\
    \midrule
    \texttt{GraDK+L3net+NLL} & $-3.303_{(0.008)}$ & $0.082$ & $0.018$ & $-2.980_{(0.003)}$ & $0.640$ & $0.079$ \\
    \texttt{GraDK+L3net+LS} & $-3.225_{(0.005)}$ & $\textbf{0.036}$ & $\textbf{0.011}$ & $-2.982_{(0.004)}$ & $\textbf{0.391}$ & $\textbf{0.067}$ \\
    \texttt{GraDK+GAT+NLL} & $-3.556_{(0.009)}$ & $0.110$ & $0.025$ & $-2.995_{(0.006)}$ & $0.942$ & $0.173$\\
    \bottomrule
    \bottomrule
  \end{tabular}
  \begin{tablenotes}
  \item *Numbers in parentheses are standard errors for three independent runs.
  \end{tablenotes}
  \end{threeparttable}
  \label{tab:real-data-results}
\end{table*}

\begin{figure}[!t]
    \centering
    \begin{subfigure}[h]{.8\linewidth}
    \includegraphics[width=1\linewidth]{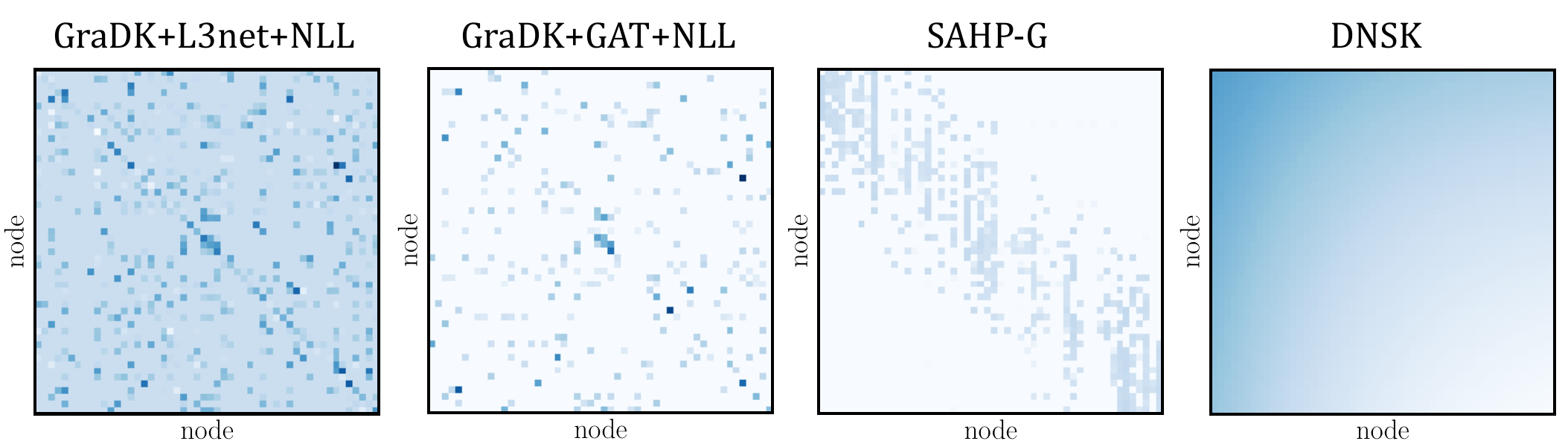}
    \end{subfigure}
\caption{Learned graph kernels for the theft data set evaluated at $t'=0$ and $t-t'=1$; the proposed method captures complex inter-node dependence compared with \texttt{DNSK} using a spatio-temporal kernel.}
\label{fig:real-data-vis}
\end{figure}

\begin{figure}[!t]
    \centering
    \begin{subfigure}[h]{.8\linewidth}
    \includegraphics[width=1\linewidth]{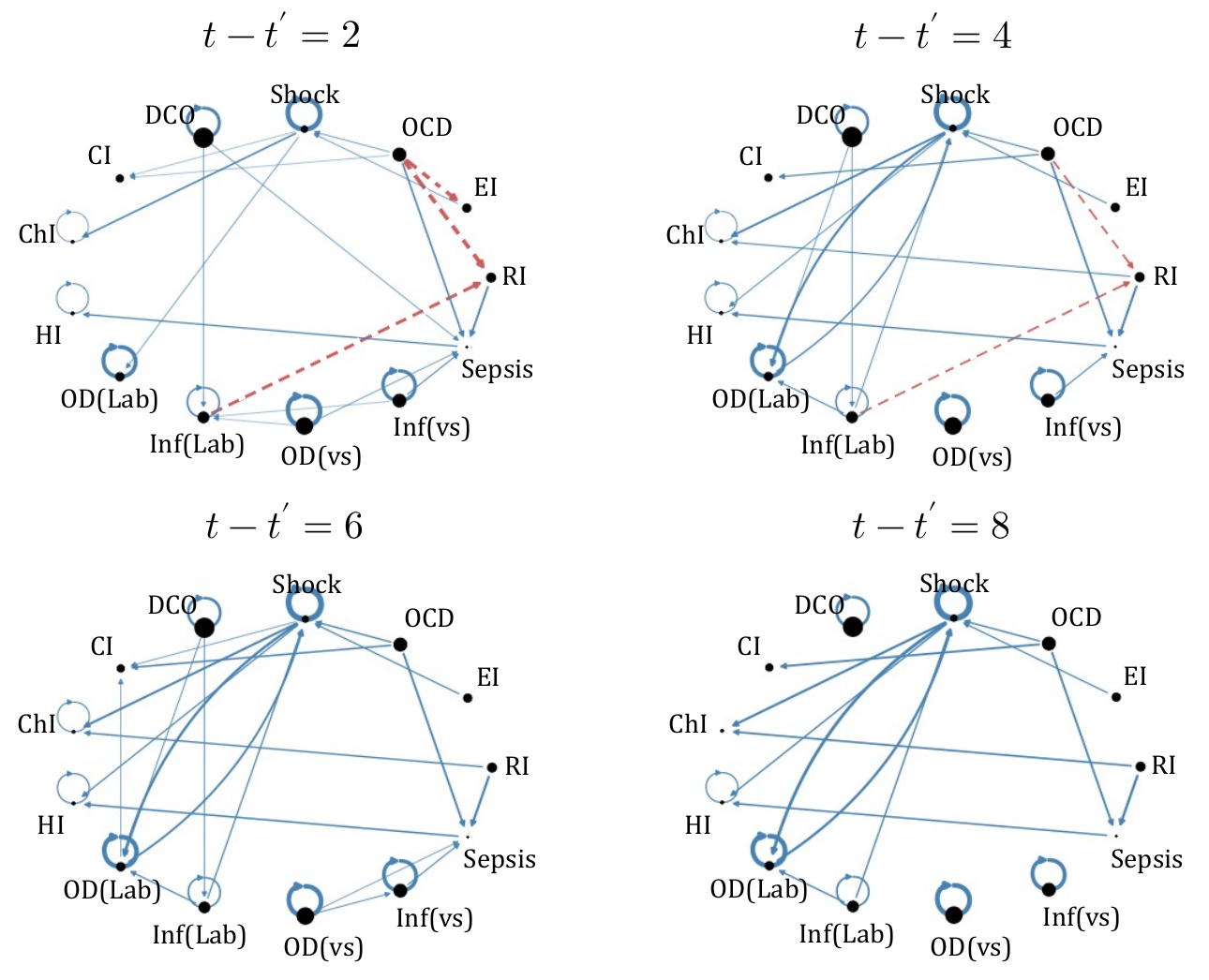}
    \end{subfigure}
    \vspace{-0.05in}
\caption{Sepsis data: Dynamic graph influence learned by \texttt{GraDK}. The learned kernel is evaluated at four time lags $2, 4, 6$, and $8$. The node radii are proportional to the background intensity on each node, and the edge widths are proportional to the influence magnitude. \revv{Solid and dashed} edges represent the excitation and inhibition effects, respectively. The history time $t'$ is set to be $0$ for all four panels.}
\label{fig:sepsis-dynamic-kernel}
\end{figure}

Table~\ref{tab:real-data-results} (and Table~\ref{tab:app-real-data-results}) show that \texttt{GraDK} achieves strong predictive performance on real datasets, as reflected by low Time MAE and Type KLD. Figure~\ref{fig:real-data-vis} further illustrates that \texttt{GraDK} recovers structured, multi-hop graph influence patterns on the theft dataset, whereas \texttt{DNSK} is less informative in this discrete graph setting and \texttt{SAHP-G} yields weaker kernel structure. Visualizations of the recovered graphs are provided in Appendix~\ref{app:additional-exp}.
\revv{Although \texttt{SAHP-G} attains slightly higher testing log-likelihood in some cases, \texttt{GraDK} provides substantially improved prediction of event times and types; this is consistent with attention models fitting local patterns more aggressively, while \texttt{GraDK} emphasizes structured multi-hop influence that improves generalization.}

\subsubsection{Dynamic graph influence on Sepsis data}

We use ICU patient time-series data for sepsis prediction to demonstrate the model’s ability to capture dynamic graph influence. Figure~\ref{fig:sepsis-dynamic-kernel} presents four snapshots of the influence patterns learned by \texttt{GraDK} among different medical indices. These snapshots are ordered by increasing time lag \( t - t' \), illustrating the temporal evolution of graph-based dependencies. The results indicate that certain medical indices (e.g., RI, OCD, DCO, OD(vs), Inf(vs)) exhibit stronger excitatory effects on sepsis onset as the onset time approaches (e.g., at \( t - t' = 2 \)). In contrast, other indices such as Shock and OD(Lab) show more interactions when sepsis does not occur. Additionally, we observe decaying inhibitory effects over time between specific index pairs, such as OCD to EI, OCD to RI, and Inf(Lab) to RI. Full names and descriptions of the medical indices are provided in Table~\ref{tab:sepsis-medical-indices}.


\section{Conclusion and discussion}

We propose a novel deep kernel framework for graph point processes that integrates a wide range of GNN architectures through localized graph filters, enabling flexible modeling of complex and non-stationary event dynamics over latent graphs. The kernel-based formulation enhances interpretability, allowing learned influence kernels to reveal dynamic interdependencies among event types. Empirically, the proposed method outperforms existing approaches in both dependency recovery and event prediction on discrete graph-structured event data, as demonstrated on synthetic and real-world datasets.

While the expressiveness of GNN-based kernels introduces additional model complexity compared to classical parametric bases, it enables the discovery of rich graph-dependent interactions directly from event data. To balance expressiveness and efficiency, our framework adopts a low-rank kernel expansion, which substantially reduces the number of trainable parameters while retaining strong empirical performance.

Several directions merit further investigation. From a theoretical perspective, it would be interesting to characterize the expressive power of GNN-based kernels for modeling spatio-temporal point processes. On the practical side, the proposed framework readily accommodates alternative GNN architectures and can be applied to a broader range of real-world event data with complex graph structures.

\section*{Acknowledgement}

YC was partially supported by NSF DMS-2134037, CMMI-2112533, and the Coca-Cola Foundation.
XC was partially supported by NSF DMS-2237842, DMS-2031849 and Simons Foundation MPS-MODL-00814643.

\bibliographystyle{apalike}
\bibliography{jcgs_refs}

\newpage
\appendix
\onecolumn

\setcounter{figure}{0} \renewcommand{\thefigure}{\thesection\arabic{figure}}
\setcounter{table}{0} \renewcommand{\thetable}{\thesection\arabic{table}}
\setcounter{equation}{0} \renewcommand{\theequation}{\thesection\arabic{equation}}
\vspace{-0.05in}

\section{Comparison with existing approaches}
\label{app:comp-with-baselines}
\vspace{-0.05in}
To elaborate on our paper's contributions further, we provide a detailed comparison between our method and existing approaches that study point processes in the context of graphs, neural networks, and graph neural networks. 
Table~\ref{tab:literature-review} compares our \texttt{GraDK} with existing methods in terms of four aspects that play crucial roles in establishing an effective and high-performing model for marked temporal point processes over graphs, which are also the main contributions of our paper that distinguish the proposed method from those in previous studies.

\begin{table}[!h]
  \caption{Comparison between our method \texttt{GraDK} with other point processes with graphs, neural networks, and graph neural networks.}
  \vspace{-0.1in}
  \centering
  \resizebox{1.\linewidth}{!}{
  \begin{threeparttable}
    \begin{tabular}{ccccc}
    \toprule
    \toprule
    Model & No parametric constraint & Modeling influence kernel & Joint modeling event time and mark on graph& Using GNN \\
    \midrule
    1. GNHP \citep{fang2023group} & \xmark & \cmark & \cmark & \xmark \\
    2. GINPP \citep{pan2023graphinformed} & \cmark & \xmark & \xmark & \xmark \\
    3. RMTPP \citep{du2016recurrent} & \xmark & \xmark & \xmark & \xmark \\
    4. NH \citep{mei2017neural} & \cmark & \xmark & \xmark & \xmark \\
    5. FullyNN \citep{omi2019fully} & \cmark & \xmark & \xmark & \xmark \\
    6. SAHP \citep{zhang2020self} & \cmark & \xmark & \xmark & \xmark \\
    7. DNSK \citep{dong2022spatio} & \cmark & \cmark & \xmark & \xmark \\
    8. LogNormMix \citep{shchur2019intensity} & \cmark & \xmark & \xmark & \xmark \\
    9. DeepSTPP \citep{zhou2022neural} & \xmark & \xmark & \xmark & \xmark \\
    10. NSTPP \citep{chen2020neural} & \cmark & \xmark & \xmark & \xmark \\
    11. DSTPP \citep{yuan2023spatio} & \cmark & \xmark & \xmark & \xmark \\
    12. THP-S \citep{zuo2020transformer} & \cmark & \xmark & \cmark & \xmark \\
    13. SAHP-G \citep{zhang2021learning} & \cmark & \xmark & \cmark & \xmark \\
    14. GHP \citep{shang2019geometric} & \xmark & \cmark & \xmark & \cmark \\
    15. GNPP \citep{xia2022graph} & \cmark & \xmark & \xmark & \cmark \\
    16. GBTPP \citep{wu2020modeling} & \xmark & \xmark & \xmark & \cmark \\
    \hline
    17. \texttt{GraDK} (Ours) & \cmark & \cmark & \cmark & \cmark \\
    \bottomrule
    \bottomrule
  \end{tabular}
  \end{threeparttable}
  }
  \label{tab:literature-review}
  \vspace{-0.05in}
\end{table}

We have compared our model with Methods 3, 5, 7, 12, and 13 in the above table by experiments and showed our advantageous performance. The comparisons between Methods 3, 5, and 7 demonstrate the benefits of considering latent graph information when modeling graph point processes. The comparisons between Methods 12 and 13 (they are deep neural point processes that incorporate known graph structures through the adjacency matrix) showcase the advantages of incorporating GNNs for flexibly capturing the complicated multi-hop, non-Euclidean event influence. 
We do not compare with (i) Methods 1 and 14, for they aim to develop theoretical results in the context of parametric point processes with graph/network structures; (ii) Method 2, since it is an extension of Method 5; (iii) Methods 4 and 8 for their similarity with Methods 3 and 5, respectively; (iv) Method 6 for it is another version of Method 13 without considering the graph structure; (v) Methods 9, 10, and 11 for they focus on spatio-temporal point processes, and we have demonstrated the advantages of our graph point process model against STPPs using ablation study in Section~\ref{sec:ablation-study}; (vi) Methods 15 and 16 for they focus on the task of temporal link prediction rather than modeling discrete marked events.

Being the first paper that \textit{incorporates GNNs in the influence kernel without any parametric constraints}, four aspects of our paper's contributions include (also highlighted in Section~\ref{sec:introduction}):

\begin{enumerate}[itemsep=3pt]
    \item We have no parametric model constraints, which becomes increasingly important in modern applications for the model to capture complex event dependencies. Prior works (\textit{e.g.}, Methods 1, 3, 14) typically assume a specific parametric form for the model, although they can be data- and computationally efficient, which may restrict the model expressiveness for modern applications with complex and large datasets. 

    \item We model the influence kernel instead of the conditional intensity, which effectively leverages the repeated influence patterns by combining statistical and deep models with excellent model interpretability.
    As shown in Table~\ref{tab:literature-review}, most previous studies fall into the category of modeling the conditional intensity function instead of the influence kernel. While such models are expressive, they ignore the benefits of leveraging the repeated influence patterns, and the model interpretability may be limited.

    \item We jointly model event time and mark over a latent graph structure, which most existing approaches fail to achieve. Some approaches (\textit{e.g.}, Methods 5, 8) enjoy great model flexibility for modeling event times, but they only focus on temporal point processes and can hardly be extended to mark space. Another body of research (\textit{e.g.}, Methods 2, 3, 8) decouples the time and mark space when modeling event distributions for model simplicity and efficiency, which sometimes limits the model flexibility and leads to sub-optimal solutions. More studies model marked events without any graph structure considered. One example is the family of existing spatio-temporal point process approaches (\textit{e.g.}, Methods 9, 10, 11) that are inapplicable in the setting of graph point processes. 

    \item We adopt GNNs in the influence kernel, which faces non-trivial challenges and requires a carefully designed model architecture. As shown in Table~\ref{tab:literature-review}, there has been limited work in exploiting underlying graph structures by leveraging GNNs. 
    Previous studies (\textit{e.g.}, Methods 2, 12, 13) combine non-graph neural network architectures (\textit{e.g.}, fully-connected neural networks, RNNs, etc.) along with certain graph information (\textit{e.g.}, adjacency matrices), rather than directly incorporating GNN structures in the models. 
    Another attempt (Method 14) integrates graph convolutional RNNs with the influence kernel. However, they focus on using GNNs to estimate the parameters of the conditional intensity function with a parametric (exponentially decaying) kernel.
    Two previous studies (Methods 15, 16) on modeling point processes resort to message-passing GNNs. However, they do not use influence kernels and focus on a different problem of graph node interaction prediction rather than modeling the dependencies and occurrences of discrete marked events.
\end{enumerate}

It is worth noting that extending the influence kernel onto the graph structure with GNN architectures incorporated is crucial and non-trivial for modeling point processes over graphs. Because (i) the lack of a straightforward definition of distance between event marks makes the distance-based spatial kernel non-sensible on a graph structure, and (ii) simply replacing the distance-based kernel with some scalar coefficients to capture the interaction between each pair of nodes will significantly limit the model expressiveness, especially in modern applications. In this paper, we design the influence kernel motivated by kernel SVD and Mercer's theorem \citep{mercer1909xvi} and introduce the concept of localized graph filter basis into the kernel to fully take advantage of the universal representation power of GNNs.

\section{Example: Graph filter bases in L3Net}\label{app:L3Net}

\begin{figure}[h]
\centering
    \begin{subfigure}[h]{.32\linewidth}
    \centering
        \includegraphics[width=\linewidth]{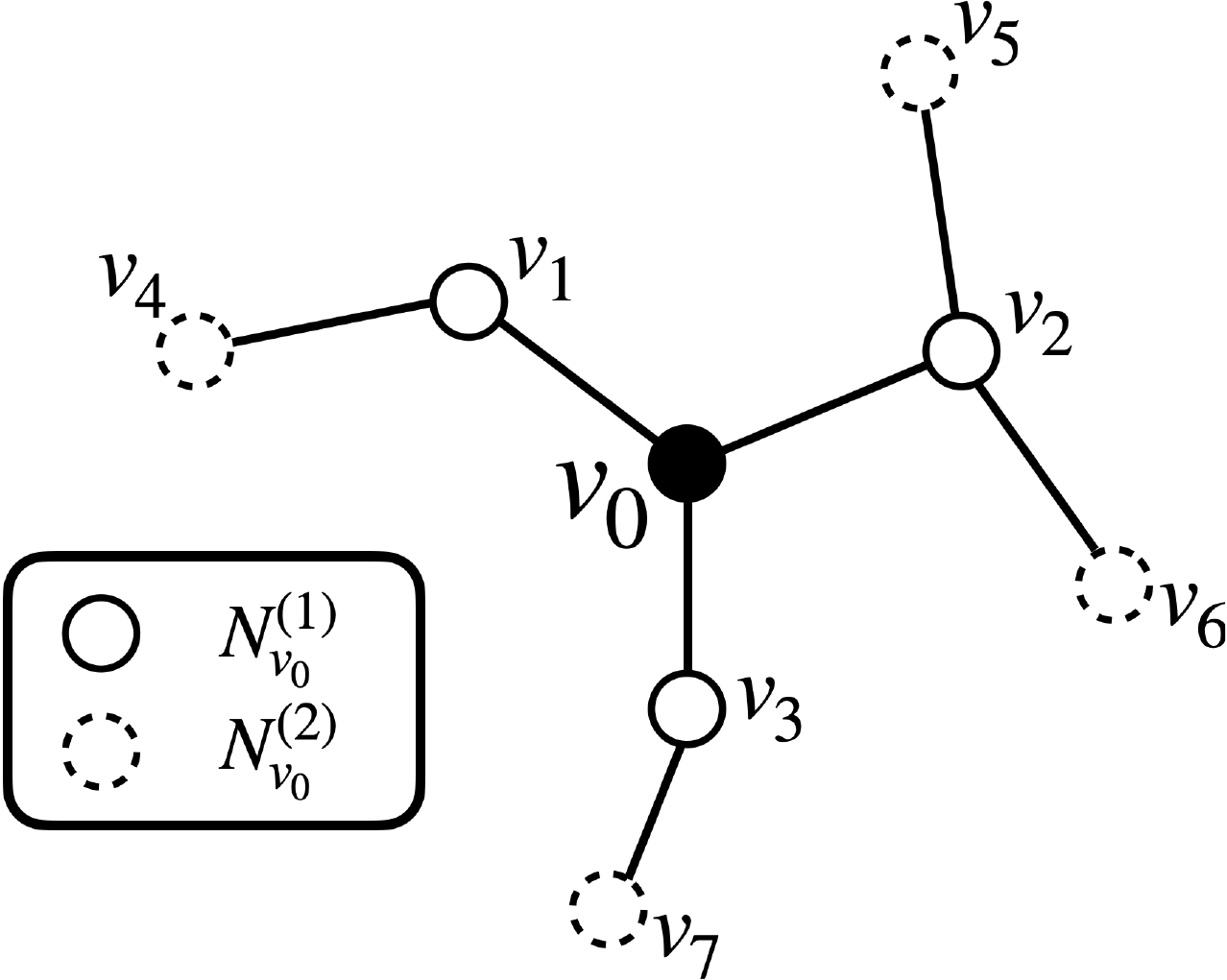}
        \caption{The latent graph structure}
    \end{subfigure}
    \hspace{0.4in}
    \begin{subfigure}[h]{.42\linewidth}
    \centering
        \includegraphics[width=\linewidth]{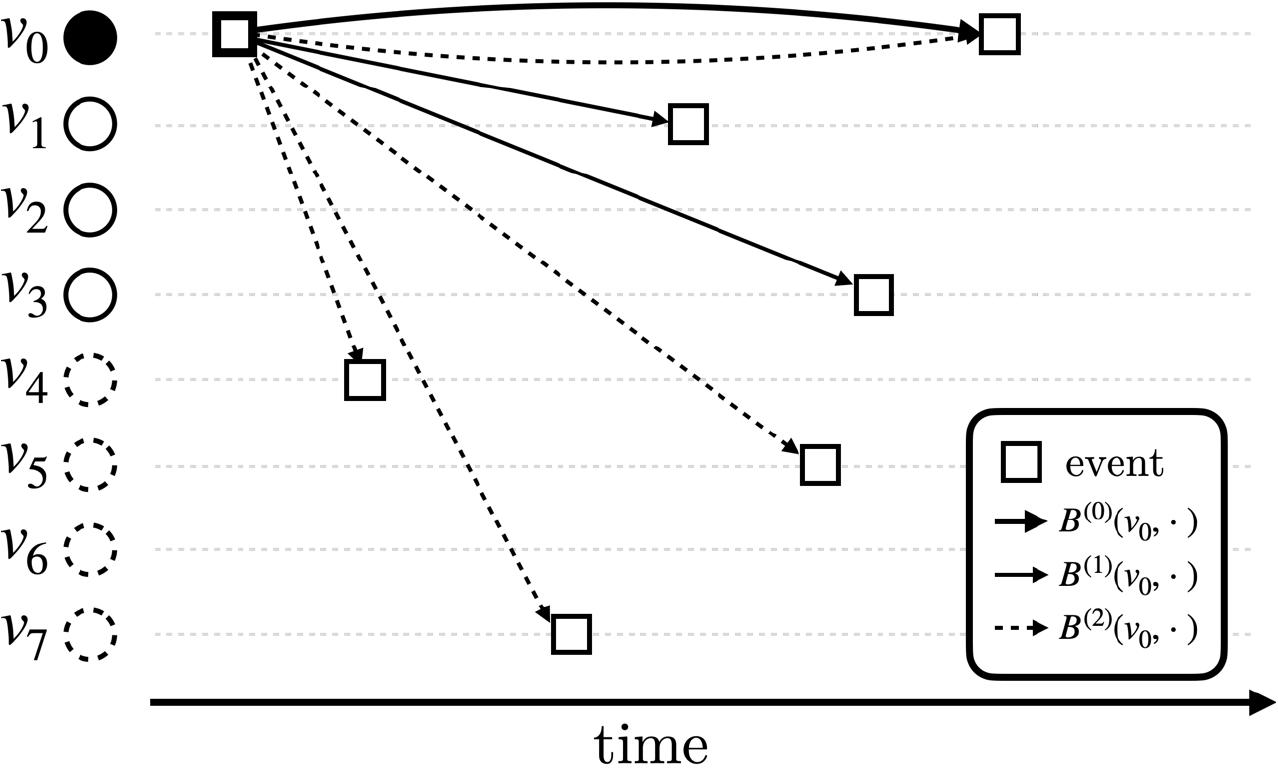}
        \caption{Events dependencies over graph nodes}
    \end{subfigure}
    \caption{An example of modeling events on an 8-node graph using graph filter bases in L3Net: (a) The latent graph structure. \revv{Hollow and dashed nodes} represent the $1$st and $2$nd order neighbors of $v_0$, denoted by $N_{v_0}^{(1)}$ and $N_{v_0}^{(2)}$, respectively.
    (b) Three graph filter bases $B^{(0)}$, $B^{(1)}$, and $B^{(2)}$ capture the dependencies between events. Hollow squares represent events observed on each node. \revv{Bold and dashed} lines indicate the potential influence of the earliest type-$v_0$ event on future events captured by different bases.
    }
    \vspace{-0.1in}
\label{fig:illustration-filter-basis}
\end{figure}

Figure~\ref{fig:illustration-filter-basis} illustrates the mechanism of the graph filter bases in L3Net to capture the event dependencies during the modeling of sequential events on an 8-node graph. The neighborhood orders are highlighted as the superscripts of each graph filter basis.

\section{Incorporation of localized graph filters in graph kernel}
\label{app:deep-graph-kernel-details}

We provide the details of incorporating different localized graph filters in graph convolutions into our proposed graph basis kernel, including the model formulation and complexity (number of trainable parameters). A graph convolution layer maps the input node feature $x$ to the output feature $y$. Particularly, for the graph convolutions in the graph neural networks discussed in Section~\ref{sec:graph-basis-kernel}, we have:

\begin{enumerate}[(1)]
    \item \textit{Chebnet} \citep{defferrard2016convolutional}: The filtering operation in Chebnet is defined as
    \vspace{-0.05in}
    \[
    \vspace{-0.05in}
        y = \sum_{r=0}^{R-1}\theta_rT_r(\tilde{L})x,
    \]
    where $T_r$ is the Chebyshev polynomial of order $r$ evaluated at the scaled and normalized graph Laplacian $\tilde{L} = 2L/\lambda_{\text{max}} - I$, where $L = I - D^{-1/2}AD^{-1/2}$, $D$ is the degree matrix, $A$ is the adjacency matrix, and $\lambda_{\text{max}}$ is the largest eigenvalue of $L$.
    
    We let $B_r = T_{r-1}(\tilde{L}) \in \mathbb{R}^{|V|\times |V|}$, which can be computed by the recurrence relation of Chebyshev polynomial $B_r = 2LB_{r-1} - B_{r-2}$.
    Since $\{B_r\}_{r=1}^{R}$ are calculated by the graph Laplacian, the trainable parameters are polynomial coefficients (in our case, the weights that combine basis kernels) of $\mathcal{O}(R)$.

    \item \textit{L3Net} \citep{cheng2020graph}: In L3Net, the graph convolution layer for single channel graph signal is specified as (omitting the bias, non-linear mapping)
    \vspace{-0.08in}
    \[
    \vspace{-0.1in}
        y = \sum_{r=1}^{R}a_r B_r^{(o_r)}(v^\prime, v)x.
    \]
    Each $B_r^{(o_r)}(v^\prime, v) \neq 0$ only if $v \in N_{v^\prime}^{(o_r)}$. 
    
    In our model, we choose an integer $o_r$ for each $B_r$ and formulate them the same way in L3Net. Thus, each $B_r(v^\prime, v)$ models the influence from $v^\prime$ to its $o_r$-th order neighbors on the graph. 
    All bases $\{B_r\}_{r=1}^{R}$ are learnable. Assuming the average number of neighbors of graph nodes is $C$, the total number of trainable parameters is of $\mathcal{O}(RC|V|)$.

    \item \textit{GAT} \citep{velickovic2017graph}: Considering GAT with $R$ attention heads, the graph convolution operator in one layer can be written as
    \vspace{-0.08in}
    \[
    \vspace{-0.1in}
        y = \sum_{r=1}^{R}\mathcal{A}^{(r)}x\Theta_r, \quad \mathcal{A}_{v^\prime, v}^{(r)} = \frac{e^{c_{v^\prime v}^{(r)}}}{\sum_{v\in N_{v^\prime}^{(1)}}e^{c_{v^\prime v}^{(r)}}}, \quad c_{v^\prime v}^{(r)} = \sigma((a^{(r)})^\top[W^{(r)}x_{v^\prime} || W^{(r)}x_{v}]).
    \]
    Here $\{a^{(r)}, W^{(r)}\}_{r=1}^{R}$ are trainable model parameters, and $\Theta_r = W^{(r)}C^{(r)}$, where $C^{(r)}$ is a fixed matrix for concatenating outputs from $R$ attention heads. Mask attention is applied to inject the graph structure, that is, $\mathcal{A}_{v^\prime, v}^{(r)} \neq 0$ only when $v \in N_{v^\prime}^{(1)}$. 
    
    Therefore, each $B_r(v^\prime, v)$ in our graph basis kernel can be a learnable localized graph filter with positive entries, and column-wise summation normalized to one to mimic the affinity matrix $\mathcal{A}^{(r)}$.
    When global attention is allowed (\textit{i.e.}, every node can attend on every other node), the number of trainable parameters is of $\mathcal{O}(R|V|^2)$. When mask attention is applied (which is commonly used in GAT), the total number of trainable parameters is $\mathcal{O}(RC|V|)$.

    \item \textit{GPS Graph Transformer} \citep{rampavsek2022recipe}: At each layer of GPS Graph Transformer, the graph node features are updated by aggregating the output of a message-passing graph neural network (MPNN) layer and a global attention layer. Assuming a sum aggregator in the MPNN layer, the filter operation can be expressed as
    \vspace{-0.08in}
    \[
    \vspace{-0.1in}
        y = \sum_{r=1}^{R} \left( \mathcal{A}^{(r)}x W_1^{(r)} + x W_2^{(r)}\right)C^{(r)}.
    \]
    Here $\mathcal{A}^{(r)}$ is the affinity matrix in the attention layer, and $W_1^{(r)}, W_2^{(r)}$ are weight matrices in the attention layer and MPNN layer, respectively. The fixed matrices $\{C^{(r)}\}$ concatenate the $R$ attention heads and MPNN layers.
    
    We can integrate such a GPS Graph Transformer structure by introducing $2R$ localized graph filter bases. For each $r$, $B_r$ takes the form of a learnable affinity matrix, and $B_{2r} = A$, where $A$ is the graph adjacency matrix.
    Here $\{B_r\}_{r=1}^{R}$ are learnable and $\{B_r\}_{r=R+1}^{2R}$ are fixed. The total number of trainable parameters with mask attention adopted is of $\mathcal{O}(RC|V|)$.
\end{enumerate}

\paragraph{Parameter complexity of the graph kernel.}
For typical localized GNN filters, the number of learnable parameters scales with graph connectivity: each graph filter contributes \( \mathcal{O}(C|V|) \) parameters, where \( C \) denotes the average neighborhood size and \( |V| \) is the number of nodes. Using a rank-\( R \) graph kernel therefore leads to a total parameter complexity of \( \mathcal{O}(RC|V|) \). In contrast, classical multivariate point process models and many neural point process approaches that ignore graph structure require \( \mathcal{O}(|V|^2) \) parameters to model all pairwise interactions. In practice, for sparse graphs and when modeling influence up to a small number of hops, we typically have \( C, R \ll |V| \), which enables efficient scaling to large graphs.

\section{Derivation of two loss functions}
\label{app:loss-derivation}

Considering one event sequence with $n$ events, we present the detailed derivation of loss functions NLL and LS in the following:

\vspace{-0.1in}
\paragraph{Negative log-likelihood (NLL).} The model likelihood of point processes can be derived from the conditional intensity \eqref{eq:conditional-intensity-with-kernel}. For the $i+1$-th event at $(t, v)$, by definition, we can re-write $\lambda(t, v)$ as follows:
\[
    \begin{aligned}
        \lambda(t, v) &= \mathbb{E}\left( \mathbb{N}_v([t, t+\Delta t] \times v)) | \mathcal{H}_{t} \right) / dt \\
        &= \mathbb{P}\{t_{i+1} \in [t, t+\Delta t], v_{i+1} = v| \mathcal{H}_{t_{i+1}} \cup \{t_{i+1} \geq t\}\} /dt\\
        &= \frac{\mathbb{P}\{t_{i+1} \in [t, t+\Delta t], v_{i+1} = v, t_{i+1} \geq t, v_{i+1} = v | \mathcal{H}_{t_{i+1}}\}/dt}{\mathbb{P}\{t_{i+1} \geq t | \mathcal{H}_{t_{i+1}}\}}. \\
        &= \frac{f(t, v)}{1 - F(t)}.
    \end{aligned}
\]
Here $\mathbb{N}_v$ is the counting measure on node $v$, $F(t) = \mathbb{P}(t_{i+1} < t | \mathcal{H}_{t_{i+1}})$ is the conditional cumulative probability and $f(t, v)$ is the corresponding conditional probability density for the next event happening at $(t, v)$ given observed history. 
We multiply the differential $dt$ on both sides of the equation and sum over all the graph nodes $V$:
\[
    \begin{aligned}
        dt \cdot \sum_{v \in V} \lambda(t, v) = \frac{dt \cdot \sum_{v \in V} f(t, v)}{1 - F(t)} = -d\log{(1 - F(t))}.
    \end{aligned}
\]
Integrating over $t$ on $[t_i, t)$ with $F(t_{i}) = 0$ leads to the fact that
\[
    F(t) = 1 - \exp \left (-\int_{t_i}^{t}\sum_{v \in V} \lambda(t, v)dt \right ),  
\]
then we have
\[
    f(t, v) = \lambda(t, v) \cdot \exp \left (-\int_{t_i}^{t}\sum_{v \in V} \lambda(t, v)dt \right ) .
\]
Using the chain rule, we have the model log-likelihood to be:
\[
    \begin{aligned}
        \ell_{\text{NLL}} &= - \log f((t_1, v_1), \dots, (t_n, v_n)) = - \log \prod_{i=1}^{n}f(t_i, v_i) \\
        &= \sum_{v\in V}\int_{0}^{T}\lambda(t, v)dt - \sum_{i=1}^{n}\lambda(t_i, v_i) .
    \end{aligned}
\]

\vspace{-0.1in}
\paragraph{Least square (LS) loss.} 
\rev{Let the counting measure of the events on graph node $v$ be denoted as $d \mathbb{N}_v (t)$, and the events are $ \{( t_i, v_i ), \, i=1,\cdots,n \}$ over time horizon $[0,T]$. 
We write the model intensity as $\lambda(t,v)$, omitting the dependence on the model parameter $\theta$ in the notation.
The (population) L2 loss takes the form
\begin{equation*}
\ell_{\rm LS} = \E \sum_{v \in V} 
    \int_0^T ( \lambda(t,v)^2 
    - 2\lambda(t,v) \lambda^*(t,v) ) dt,
\end{equation*}
which, up to a constant independent from the model, equals $\E \sum_{v \in V} 
    \int_0^T (\lambda(t,v) 
    -  \lambda^*(t,v))^2 dt$.
By definition, $\lambda^*(t,v) dt = \E[ d \mathbb{N}_v (t) | \calH_t ]$. Together with that $\lambda(t,v) \in \calH_t$, we have
\begin{align*}
\ell_{\rm LS} 
& = \E \sum_{v \in V} 
    \int_0^T \left( \lambda(t,v)^2 dt 
    - 2\lambda(t,v) \E[ d \mathbb{N}_v (t) | \calH_t ] \right) \\
& =  \E \sum_{v \in V} 
    \int_0^T \left( \lambda(t,v)^2 
    - 2\lambda(t,v)  d \mathbb{N}_v (t) \right)  \\
& =  \E \sum_{v \in V} 
    \int_0^T \lambda(t,v)^2 
    - 2 \sum_{i=1}^n \lambda(t_i,v_i).  
\end{align*}
Replacing the $\E$ by the sample average over event trajectories obtain the empirical LS loss over the training event data. 
}



\section{Efficient model computation}
\vspace{-0.05in}
\label{app:efficient-computation}




In our optimization problem, the log-barrier term $p(\theta, b)$ guarantees the validity of the learned models by penalizing the value of the conditional intensity function on a dense enough grid over space, denoted as $\mathcal{U}_{\text{bar}, t} \times V$ where $ \mathcal{U}_{\text{bar}, t} \subset [0, T]$. 
The optimization problems with two loss functions are expressed as 
\begin{equation}
    \begin{aligned}
        & \min _{\theta} \mathcal{L}_{\text{NLL}}(\theta) := \\
        & \frac{1}{M}\sum_{s=1}^{M} \left( \sum_{v \in V}\int_0^T\lambda(t, v)dt - \sum_{i=1}^{n_s} \log \lambda\left(t_i^s, v_i^s\right)
    \right ) - \frac{1}{wM|\mathcal{U}_{\text{bar}, t} \times V|}\sum_{s=1}^{M}\sum_{t \in \mathcal{U}_{\text{bar}, t}}\sum_{v \in V}\log(\lambda(t, v) - b),
    \end{aligned}
    \label{eq:mle-optimization}
\end{equation}
and
\begin{equation}
    \begin{aligned}
        & \min _{\theta} \mathcal{L}_{\text{LS}}(\theta) := \\
        & \frac{1}{M}\sum_{s=1}^{M} \left(\sum_{v \in V}\int_{0}^{T}\lambda^2(t, v)dt - \sum_{i=1}^{n_s}2\lambda(t_i^s, v_i^s)\right) - \frac{1}{wM|\mathcal{U}_{\text{bar}, t} \times V|}\sum_{s=1}^{M}\sum_{t \in \mathcal{U}_{\text{bar}, t}}\sum_{v \in V}\log(\lambda(t, v) - b),
    \end{aligned}
    \label{eq:l2-optimization}
\end{equation}
respectively.
The computational complexity associated with calculating the objective function has consistently posed a limitation for neural point processes. The evaluations of neural networks are computationally demanding, and traditional numerical computation requires model evaluations between every pair of events. Consider one event sequence $\{(t_i, v_i)\}_{i=1}^{n}$ in $\mathcal{S}$ with a total number of $n$ events; traditional methods have a complexity of $\mathcal{O}(n^2)$ for neural network evaluation. In the following, we showcase our efficient model computation of complexity $\mathcal{O}(n)$ for two objective functions using a domain discretization strategy.

\vspace{-0.1in}
\paragraph{Negative log-likelihood.}
We identify three distinct components in the optimization objective \eqref{eq:mle-optimization} as log-summation, integral, and log-barrier, respectively. We introduce a uniform grid $\mathcal{U}_t$ over time horizon $[0, T]$, and the computation for each term can be carried out as follows:

\begin{itemize}
    \item \textit{Log-summation}. We plug in the influence kernel \eqref{eq:kernel-formulation} to the log-summation term and have
    \[
    \sum_{i=1}^{n}\log\lambda(t_i, v_i) = \sum_{i=1}^{n}\log{\left(\mu + \sum_{t_j < t_i}\sum_{r=1}^{R}\sum_{l=1}^{L}\alpha_{lr} \psi_l(t_j)\varphi_l(t_i - t_j)B_r(v_j, v_i)\right)}.
    \]
    Each $\psi_l$ is only evaluated at event times $\{t_i\}_{i=1}^{n}$. To prevent redundant evaluations of the function $\varphi_l$ for every pair of event times $(t_i, t_j)$, we restrict the evaluation of $\varphi_l$ on the grid $\mathcal{U}_t$. By adopting linear interpolation between two neighboring grid points of $t_i - t_j$, we can determine the value of $\varphi_l(t_i - t_j)$. In practice, the influence of past events is limited to a finite range, which can be governed by a hyperparameter $\tau_{\text{max}}$. Consequently, we can let $\varphi_l(\cdot)=0$ when $t_i - t_j > \tau_{\text{max}}$ without any neural network evaluation. The computation of $B_r(v_j, v_i)$ is accomplished using matrix indexing, a process that exhibits constant computational complexity when compared to the evaluation of neural networks.

    \item \textit{Integral}. The efficient computation of the integral benefits from the formulation of our conditional intensity function. We have
    \begin{align*}
    \sum_{v \in V}\int_{0}^{T}\lambda(t, v)dt & = \mu |V|T + \sum_{i=1}^{n}\sum_{v \in V}\int_{0}^{T}I(t_i < t)k(t_i, t, v_i, v)dt \\ 
    & = \mu |V| T + \sum_{i=1}^{n}\sum_{r=1}^{R}\sum_{v=1}^{V}B_r(v_i, v)\sum_{l=1}^{L}\alpha_{rl}\psi_l(t_i)\int_{0}^{T-t_i}\varphi_l(t)dt.
    \end{align*}
    Similarly, the evaluations of $B_r(v_i, v)$ are the extractions of corresponding entries in the matrices, and $\psi_l$ is only evaluated at event times $\{t_i\}_{i=1}^{n}$. We leverage the evaluations of $\varphi_l$ on the grid $\mathcal{U}_t$ to facilitate the computation of the integral of $\varphi_l$. Let $F_l(t) := \int_{0}^{t}\varphi_l(\tau)d\tau$. The value of $F_l$ on $j$-th grid point in $\mathcal{U}_t$ equals the cumulative sum of $\varphi_l$ from the first grid point up to the $j$-th point, multiplied by the grid width. Then, $F_l(T-t_i)$ can be computed by the linear interpolation of values of $F_l$ at two adjacent grid points of $T - t_i$.

    \item \textit{Log-barrier}. The computation can be carried out similarly to the computation of the log-summation term by replacing $(t_i, v_i)$ with $(t, v)$ where $t \in \mathcal{U}_{\text{bar}, t}, v \in V$.
\end{itemize}

\vspace{-0.1in}
\paragraph{Least square loss.} Similarly, we term the three components in the objective function \eqref{eq:l2-optimization} as square integral, summation, and log-barrier, respectively. The terms of summation and log-barrier are calculated in the same way as the log-summation and log-barrier in \eqref{eq:mle-optimization}, respectively, except for no logarithm in the summation. Since expanding the integral after squaring the conditional intensity function is complicated, we leverage the evaluations of the intensity function on the dense enough grid for log-barrier penalty $\mathcal{U}_{\text{bar}, t} \times V$ and use numerical integration for computing the square integral.

\begin{table}[!t]
  \caption{Comparison of the computation time for two loss functions on each synthetic data set.}
  \centering
  \resizebox{.9\linewidth}{!}{
  \begin{threeparttable}
    \begin{tabular}{cccccc}
    \toprule
    \toprule
    Model & {\shortstack{3-node graph \\ with negative influence}} & {\shortstack{16-node graph \\ with 2-hop influence}} & {\shortstack{50-node graph}} & {\shortstack{50-node graph \\ with non-separable kernel}} & {\shortstack{225-node graph}}\\  
    \midrule
    \texttt{GraDK+NLL} & $0.931$ & $3.993$ & $25.449$ & $30.348$ & $4.948$\\
    \texttt{GraDK+LS} & $0.985$ & $3.927$ & $26.182$ & $29.912$ & $5.619$\\
    \bottomrule
    \bottomrule
  \end{tabular}
  \begin{tablenotes}
  \item *Unit: seconds per epoch.
  \end{tablenotes}
  \end{threeparttable}
  }
  \label{tab:running-time-comparison}
\end{table}

\vspace{0.2in}
\noindent We provide the analysis of the $\mathcal{O}(n)$ computational complexity for both objective functions. For objective \eqref{eq:mle-optimization}, 
the evaluation of $\{\psi_l\}_{l=1}^{L}$ has a complexity of $\mathcal{O}(Ln)$, while the evaluation of $\{\varphi_l\}_{l=1}^{L}$ requires $\mathcal{O}(L|\mathcal{U}_t|)$ complexity. On the other hand, $\{B_r\}_{r=1}^{R}$ are computed with $\mathcal{O}(1)$ complexity. Therefore, the total complexity of our model computation is $\mathcal{O}(n + |\mathcal{U}_t|)$. Moreover, the grid selection in practice is flexible, striking a balance between learning accuracy and efficiency. For instance, the number of grid points in $\mathcal{U}_t$ and $\mathcal{U}_{\text{bar}, t}$ can be both chosen around $\mathcal{O}(n)$, leading to an overall computational complexity of $\mathcal{O}(n)$. Similar analysis can be carried out for the objective \eqref{eq:l2-optimization}. It is worth noting that all the evaluations of neural networks and localized filter bases can be implemented beforehand, and both optimization objectives are efficiently calculated using basic numerical operations. Table~\ref{tab:running-time-comparison} shows the wall clock times for model training with each loss function on all the synthetic data sets, indicating that the computation time for NLL and LS are similar.

\section{Algorithm}
\label{app:algorithms}

The weight $w$ and the lower bound $b$ need to be adjusted accordingly during optimization to learn the valid model. For example, non-sensible solutions (negative intensity) appear frequently at the early stage of the training process, thus the lower bound $b$ should be close enough to the minimum of all the evaluations on the grid to effectively steer the intensity functions towards non-negative values, and the weight $w$ can be set as a small value to magnify the penalty influence. When the solutions successfully reach the neighborhood of the ground truth and start to converge (thus, no negative intensity would appear with a proper learning rate), the $b$ should be away from the intensity evaluations (\textit{e.g.}, upper-bounded by $0$), and $w$ should be large enough to remove the effect of the log-barrier penalty. We provide our training algorithm using stochastic gradient descent in the following.

\begin{algorithm}[!ht]
\begin{algorithmic}
    \STATE {\bfseries Input}: Training set $X$, batch size $M$, epoch number $E$, learning rate $\gamma$, constant $a > 1$ to update $w$ in \eqref{eq:mle-optimization} or \eqref{eq:l2-optimization}. \;
    \STATE {\bfseries Initialization:} model parameter $\theta_0$, first epoch $e=0$, $s = s_0$. Conditional intensity lower bound $b_0$.\; 
    \WHILE{$e < E$}
        \STATE Set $b_{temp} = 0$.\;
        \FOR{each batch with size $M$}
            \STATE 1. Compute $\ell(\theta)$, $\{\lambda(t_{c_t}, s_{c_s})\}_{c_t=1, \dots, |\mathcal{U}_{\text{bar}, t}|, {c_s}=1, \dots, |V|}$. \;
            \vspace{.05in}
            \STATE 2. Compute $\mathcal{L}(\theta) = -\ell(\theta) + \frac{1}{w}p(\theta, b_e)$. \;
            \vspace{.05in}
            \STATE 3. Update $\theta_{e+1} \leftarrow \theta_{e} - \gamma\frac{\partial
            \mathcal{L}}{\partial \theta_{e}}$. \;
            \vspace{.05in}
            \STATE 4. Set $b_{temp} = \min\left\{ \min\{\{\lambda(t_{c_t}, s_{c_s})\}_{c_t=1, \dots, |\mathcal{U}_{\text{bar}, t}|, c_s=1, \dots, |\mathcal{U}_{\text{bar}, s}|} - \epsilon, b_{temp}\right\}$, where $\epsilon$ is a small value to guarantee logarithm feasibility. \;
            \vspace{.05in}
        \ENDFOR
        
        $e \leftarrow e + 1, w \leftarrow w \cdot a, b_e = b_{temp}$
    \ENDWHILE
\end{algorithmic}
\caption{Model learning}
\label{alg:log-barrier-learning-algorithm}
\end{algorithm}

\section{Choice of kernel rank}
\label{app:hyperparameter}

\revv{This appendix provides empirical guidance for selecting the kernel ranks $L$ and $R$ 
introduced in Section}~\ref{sec:graph-basis-kernel}.
The kernel rank can be chosen using two approaches. The first approach treats the kernel rank as a model hyperparameter, and its selection is carried out through cross-validation. This process aims to optimize the log-likelihood on validation datasets. Each choice of $(L, R)\in[1,2]\times[1,2,3]$ has been assessed for \texttt{GraDK+L3net+NLL} on synthetic and real-world datasets. The order of the $r$-th filter is set to be $r$ (Tables~\ref{tab:LR-testing-llk-results} and \ref{tab:LR-MAE/KLD-results}). The log-likelihood values in the synthetic data show only minor differences given different hyperparameter choices once the parameters are chosen such that they sufficiently capture temporal or multi-hop graph influence (typically requiring $R>1$). In the real-world data, $L=1$ with $R=3$ yields the highest log-likelihood on the validation dataset, which are the parameters used in Section~\ref{sec:real_data_experiment}.

\paragraph{Effective rank and coefficient decay.}
Beyond cross-validation, the kernel rank can also be interpreted as a data-driven measure of model complexity. Under standard regularity assumptions for kernel SVD, the singular values of the kernel decay toward zero, motivating a low-rank approximation. In our parameterization, the magnitudes of the learned coefficients \( \alpha_{rl} \) reflect the effective contribution of each temporal–graph basis pair. As a result, the effective kernel rank can be determined by retaining significant coefficients while discarding higher-order ones, without the need to explicitly select \( D \). Empirically, we find that low-rank kernels (typically \( R=2 \)–\( 3 \)) are sufficient once multi-hop graph influence is captured; larger values of \( R \) may be used in more complex settings when computational resources permit.

\begin{figure}[!t]
    \centering
    \begin{subfigure}[h]{.78\linewidth}
    {\includegraphics[width=\linewidth]{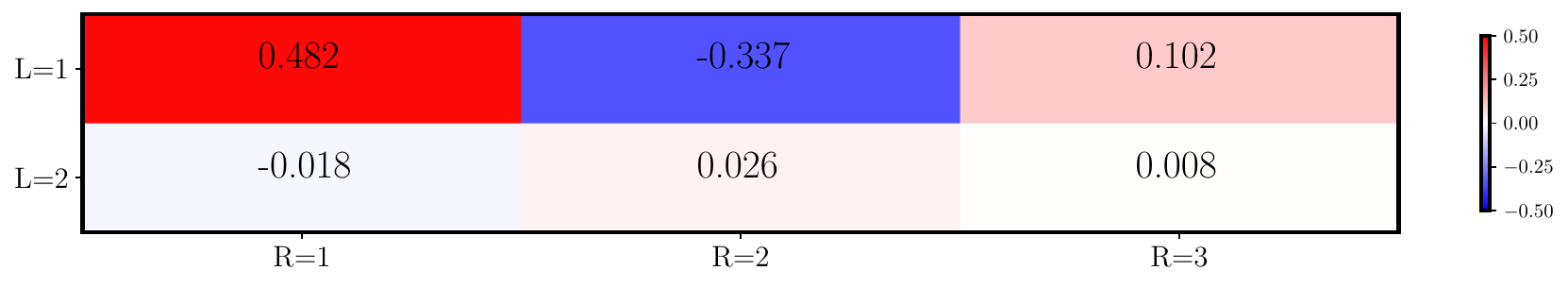}}
    \end{subfigure}
\caption{Learned $\{\alpha_{rl}\}$ of \texttt{GraDK} with $L=2, R=3$ on wildfire data.}
\label{fig:LR-wildfire-vis}
\end{figure}

The rank of the kernel can also be viewed as a level of model complexity to be learned from the data. Generally, under certain regularity assumptions of the kernel in kernel SVD, the singular values will decay towards 0, resulting in a low-rank kernel approximation. The singular values are captured by the coefficients $\alpha_{rl}$. The kernel rank is chosen by preserving significant coefficients while discarding higher-order ones; we can treat each $\alpha_{rl}$ as one learnable parameter without the need to choose $D$. The effectiveness of this approach is showcased by the results of the wildfire data. We set $L=2$ and $R=3$ at first, and then fit the model, and visualize the learned kernel coefficients in Figure~\ref{fig:LR-wildfire-vis}, which suggests that $L=1$ and $R=3$ would be an optimal choice for the kernel rank. This choice is used in subsequent model fitting.

\paragraph{Cross-validation with temporally dependent point process data.}

\revv{
Model tuning in Appendix~G is performed using validation log-likelihood evaluated on held-out samples, with splitting conducted at the trajectory or subsequence level rather than at the individual event level. For synthetic experiments, we generate multiple independent realizations of the point process over a fixed time horizon and treat each realization as one sample; training and validation sets are formed by splitting these independent trajectories. For real datasets consisting of long event sequences, we partition the data into non-overlapping subsequences of equal time duration, each treated as one sample, and perform validation by holding out entire subsequences.}

\revv{Since real-world event data may exhibit temporal dependence across time, adjacent subsequences may not be strictly independent. To mitigate this effect, we choose subsequence lengths sufficiently long so that subsequences that are well separated in time are approximately independent, relying on the mixing behavior commonly assumed for stochastic processes. This block-based validation strategy is a standard approximation for handling temporal dependence in time series and point process models. Importantly, no events from the same trajectory or subsequence appear in both the training and validation sets, ensuring that validation likelihood is computed on fully held-out data without temporal information leakage.
}

\begin{table}[!t]
  \caption{Testing log-likelihood for \texttt{GraDK} with different kernel ranks.}
  \vspace{-0.1in}
  \centering
  \resizebox{1.\linewidth}{!}{
  \begin{threeparttable}
    \begin{tabular}{ccccccccccccc}
    \toprule
    \toprule
    & \multicolumn{3}{c}{16-node graph} & \multicolumn{3}{c}{50-node graph} & \multicolumn{3}{c}{Wildfire} & \multicolumn{3}{c}{Theft} \\
    \cmidrule(lr){2-4} \cmidrule(lr){5-7}  \cmidrule(lr){8-10} \cmidrule(lr){11-13}
    Kernel rank & $R=1$ & $R=2$ & $R=3$ & $R=1$ & $R=2$ & $R=3$ & $R=1$ & $R=2$ & $R=3$ & $R=1$ & $R=2$ & $R=3$\\
    \midrule
    $L=1$ & $-3.032_{(<0.001)}$ & $-2.997_{(0.001)}$ & \colorbox{gray!20}{$-2.995_{(0.002)}$} & $-1.085_{(<0.001)}$ & \colorbox{gray!20}{$\textbf{--1.065}_{(0.002)}$} & $-1.067_{(0.001)}$ & $-3.900_{(0.004)}$ & $-3.654_{(0.009)}$ & \colorbox{gray!20}{$\textbf{--3.625}_{(0.002)}$} & $-3.405_{(0.001)}$ & $-3.039_{(0.011)}$ & \colorbox{gray!20}{$\textbf{--2.980}_{(0.003)}$}\\
    $L=2$ & $-3.030_{(0.001)}$ & $-2.992_{(0.002)}$ & {$\textbf{--2.987}_{(0.005)}$} & $-1.085_{(0.001)}$ & {$-1.067_{(0.002)}$} & $\textbf{--1.065}_{(0.002)}$ & $-3.977_{(0.005)}$ & $-3.657_{(0.007)}$ & $-3.649_{(0.017)}$ & $-3.396_{(0.008)}$ & $-3.047_{(0.026)}$ & $-3.355_{(0.034)}$ \\
    \bottomrule
    \bottomrule
  \end{tabular}
  \begin{tablenotes}
  \item *Numbers in parentheses are standard errors for three independent runs. Grey boxes indicate the choice in the original paper.
  \end{tablenotes}
  \end{threeparttable}
  }
  \label{tab:LR-testing-llk-results}
\end{table}

\begin{table}[!t]
  \caption{Predictive metrics (Time MAE and Type KLD) for \texttt{GraDK} with different kernel ranks.}
  \vspace{-0.1in}
  \centering
  \resizebox{1.\linewidth}{!}{
  \begin{threeparttable}
    \begin{tabular}{ccccccccccccc}
    \toprule
    \toprule
    & \multicolumn{3}{c}{16-node graph} & \multicolumn{3}{c}{50-node graph} & \multicolumn{3}{c}{Wildfire} & \multicolumn{3}{c}{Theft} \\
    \cmidrule(lr){2-4} \cmidrule(lr){5-7}  \cmidrule(lr){8-10} \cmidrule(lr){11-13}
    Kernel rank & $R=1$ & $R=2$ & $R=3$ & $R=1$ & $R=2$ & $R=3$ & $R=1$ & $R=2$ & $R=3$ & $R=1$ & $R=2$ & $R=3$\\
    \midrule
    $L=1$ & $0.272/0.003$ & $0.153/0.001$ & \colorbox{gray!20}{$\textbf{0.028/<0.001}$} & $1.912/0.004$ & \colorbox{gray!20}{$\textbf{1.023/0.004}$} & $3.118/0.006$ & $0.943/0.107$ & $0.358/0.024$ & \colorbox{gray!20}{$\textbf{0.207/0.006}$} & $2.787/0.097$ & $1.052/0.082$ & \colorbox{gray!20}{$\textbf{0.640/0.079}$}\\
    
    $L=2$ & $0.226/0.002$ & $0.145/0.001$ & $0.064/0.002$ & $3.698/0.006$ & $1.024/0.004$ & $5.217/0.013$ & $0.945/0.109$ & $0.314/0.020$ & $0.946/0.110$ & $2.702/0.091$ & $1.001/0.082$ & $1.123/0.109$ \\
    \bottomrule
    \bottomrule
  \end{tabular}
  \begin{tablenotes}
  \item *Grey boxes indicate the choice in the original paper.
  \end{tablenotes}
  \end{threeparttable}
  }
  \label{tab:LR-MAE/KLD-results}
\end{table}

\section{Experimental details and additional results}
\label{app:additional-exp}

In this section, we give details regarding the experiments in Section~\ref{sec:experiments}, including a description of the ground truth point process models for synthetic data, latent graph structures for real data, experimental setup, and additional results. \revv{The code used to generate all experimental results in this paper is available at \url{https://github.com/xieyao14/graph-kernel}, together with instructions for reproducing the experiments.} \revv{We begin with a list of abbreviations.}

\begin{center}
\small
\begin{tabular}{ll}
\hline
\textbf{Abbreviation} & \textbf{Description} \\
\hline
GNN & Graph Neural Network \\
GAT & Graph Attention Network \\
L3Net & Low-rank Learnable Local Graph Network \\
GPS & General, Powerful, Scalable Graph Transformer \\
MHP & Multivariate Hawkes Process \\
ETAS & Epidemic-Type Aftershock Sequence \\
SAHP-G & Graph Self-Attentive Hawkes Process \\
THP-S & Structured Transformer Hawkes Process \\
DNSK & Deep Non-Stationary Kernel \\
STPP & Spatio-Temporal Point Process \\
NLL & Negative Log-Likelihood \\
LS & Least Squares \\
MAE & Mean Absolute Error \\
KLD & Kullback--Leibler Divergence \\
SAD & Sepsis-Associated Derangement \\
\hline
\end{tabular}
\end{center}

\subsection{Data description}

Our experiments are implemented using five synthetic and four real-world data sets. 

\begin{figure}[!t]
    \centering
    \begin{subfigure}[h]{.24\linewidth}
    \includegraphics[width=1\linewidth]{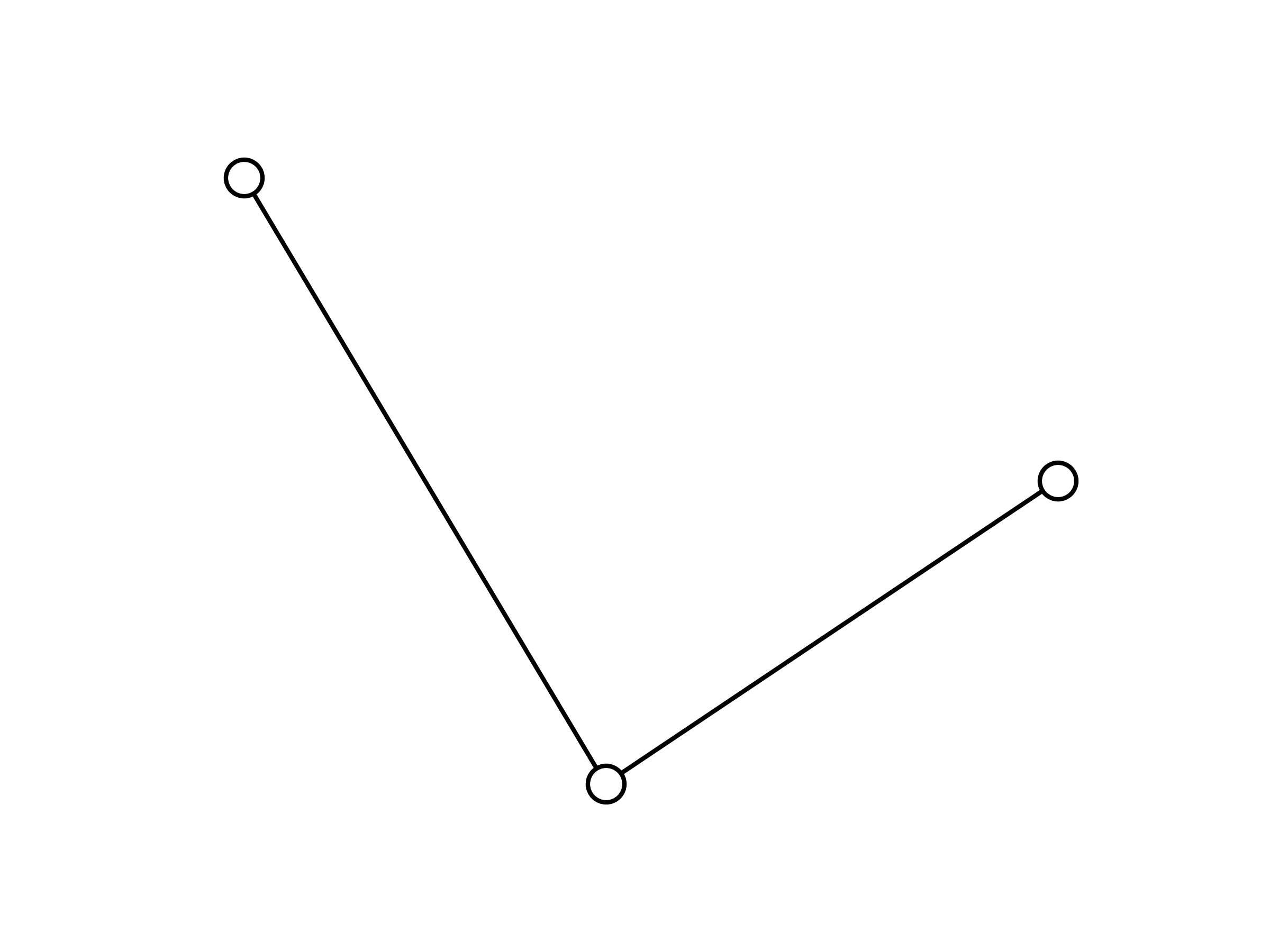}
    \caption{3-node graph}
    \end{subfigure}
    \begin{subfigure}[h]{.24\linewidth}
    \includegraphics[width=1\linewidth]{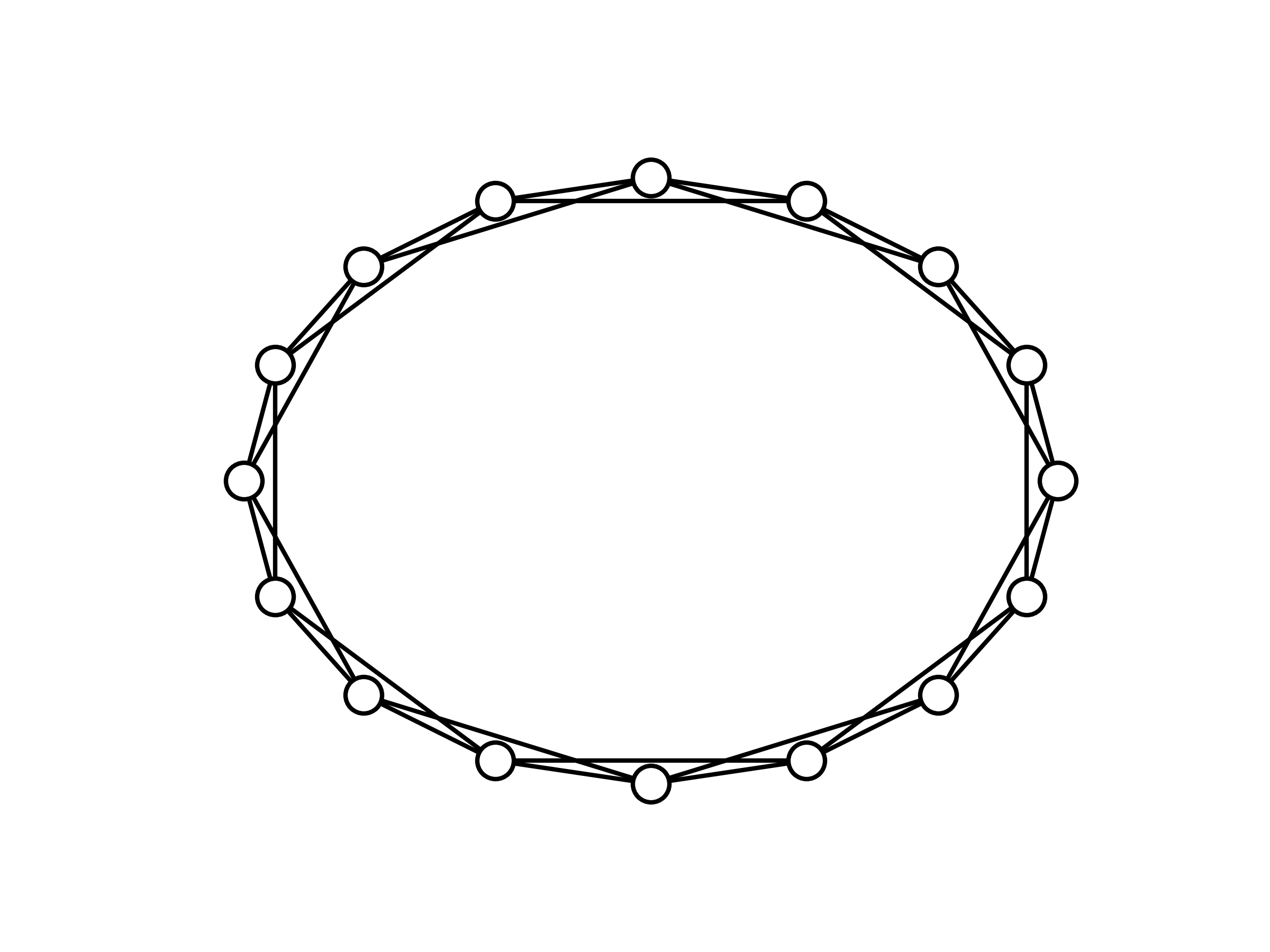}
    \caption{16-node graph}
    \end{subfigure}
    \begin{subfigure}[h]{.24\linewidth}
    \includegraphics[width=1\linewidth]{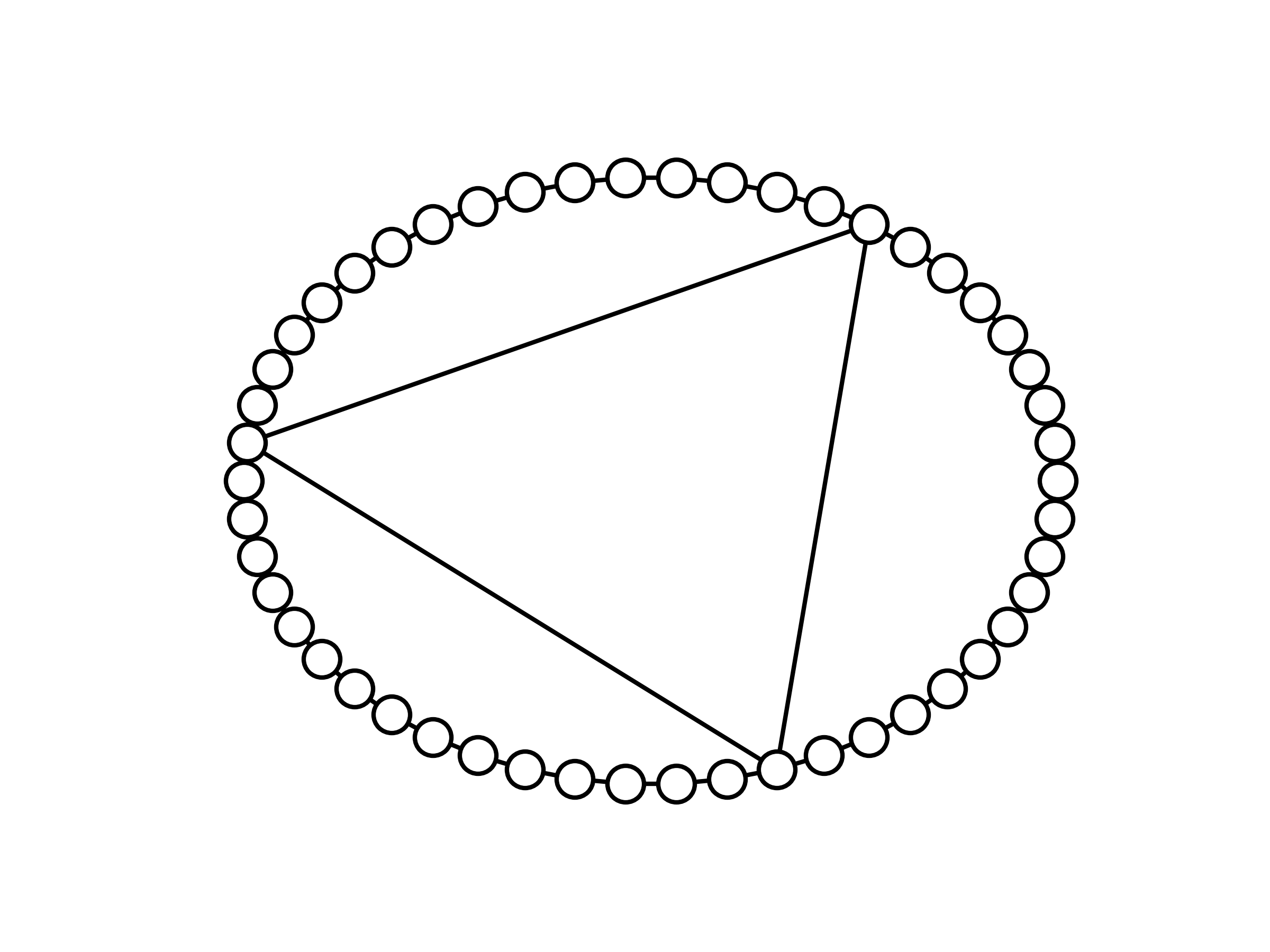}
    \caption{50-node graph}
    \end{subfigure}
    \begin{subfigure}[h]{.24\linewidth}
    \includegraphics[width=1\linewidth]{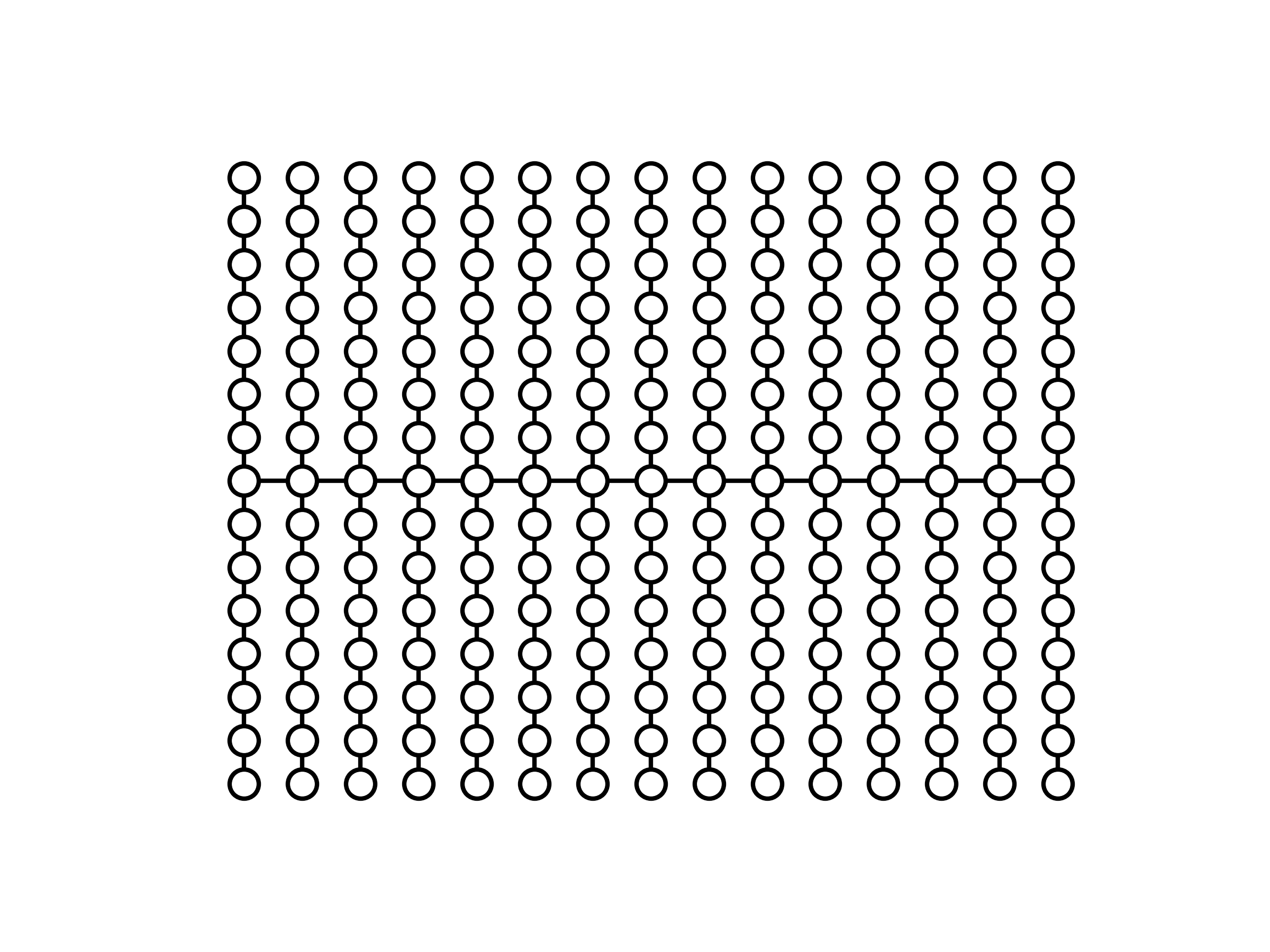}
    \caption{225-node graph}
    \end{subfigure}
\caption{Latent graph structures for the synthetic data sets. From left to right, the graphs correspond to the ones in the 3-node graph (synthetic data 1), the 16-node ring graph (synthetic data 2), the 50-node graph (synthetic data 3 \& 4), and the 225-node graph (synthetic data 5), respectively.}
\label{fig:synthetic-data-graphs}
\end{figure}

\paragraph{Synthetic data.} We provide a detailed description of the ground truth kernels and latent graph structures we used for synthetic data generation:

\begin{itemize}
    
    \item[(i)] 3 nodes with a non-stationary temporal kernel and 1-hop (positive and negative) graph influence:
    $$
    k(t^\prime, t, v^\prime, v) = 1.5(0.5+0.5\cos(0.2t^\prime))e^{-2(t-t^\prime)} \left(0.5 B_1^{(0)}(v^\prime, v) + 0.2 B_2^{(1)}(v^\prime, v) \right),
    $$
    where $B_1^{(0)} = \text{diag}(0.5, 0.7, 0.5)$, $B_2^{(1)}(2,1) = -0.2$, and $B_2^{(1)}(2,3) = 0.4$. The true graph kernel is visualized in the first panel of Figure~\ref{fig:3-node-synthetic-data-vis}.
    
    \item[(ii)] 16 nodes with a non-stationary temporal kernel and 2-hop graph influence: 
    $$
    k(t^\prime, t, v^\prime, v) = 1.5(0.5+0.5\cos(0.2t^\prime))e^{-2(t-t^\prime)} \left(0.2 B_{1}^{(0)}(v^\prime, v) - 0.3 B_{2}^{(1)}(v^\prime, v) + 0.1 B_{3}^{(2)}(v^\prime, v) \right),
    $$
    such that $(0.2 B_{1}^{(0)} - 0.3 B_{2}^{(1)} + 0.1 B_{3}^{(2)})=(0.2 I - 0.3 \tilde{L} + 0.1 (2\tilde{L}^2 - I))$. Here $\tilde{L}$ is the scaled and normalized graph Laplacian defined in Appendix~\ref{app:deep-graph-kernel-details}. This graph influence kernel is visualized in the top row of Figure~\ref{fig:16-node-synthetic-data-vis} (first column).
    
    \item[(iii)] 50 nodes with an exponentially decaying temporal kernel: $g(t^\prime, t) = 2e^{-2(t-t^\prime)}$. The graph kernel is constructed such that the influence follows a Gaussian density (with 3 modes) along the diagonal of the graph influence kernel with random noise in addition to off-ring influence. The true graph kernel is visualized in the first panel of Figure~\ref{fig:50-node-synthetic-data-vis}.

     \item[(iv)] 50 nodes with a non-separable diffusion graph kernel: $k(t^\prime, t, v^\prime, v) = e^{-2(t-t^\prime)} \cdot \frac{1}{8\pi(t-t')}e^{-\frac{d^2(v, v')}{8(t-t')}}$. The true graph kernel is visualized in the first panel of Figure~\ref{fig:non-separable-kernel-synthetic-data-vis}.

    \item[(v)] 225 nodes with exponentially decaying temporal kernel: $g(t^\prime, t) = 2e^{-2(t-t^\prime)}$. The true graph kernel is visualized in the first panel of Figure~\ref{fig:225-node-synthetic-data-vis}.
    
\end{itemize}
The latent graph structures for these five synthetic data experiments can be found in Figure~\ref{fig:synthetic-data-graphs}. Data sets are simulated using the thinning algorithm \citep{daley2007introduction}. \revv{The time window length is $T=50$ for datasets (i)--(ii) and $T=10$ for datasets (iii)--(v).} Each data set contains 1,000 sequences with an average length of 50.9, 105.8, 386.8, 558.1, and 498.3, respectively.

\begin{figure}[!t]
    \centering
    \begin{subfigure}[h]{.4\linewidth}
    \includegraphics[width=1\linewidth]{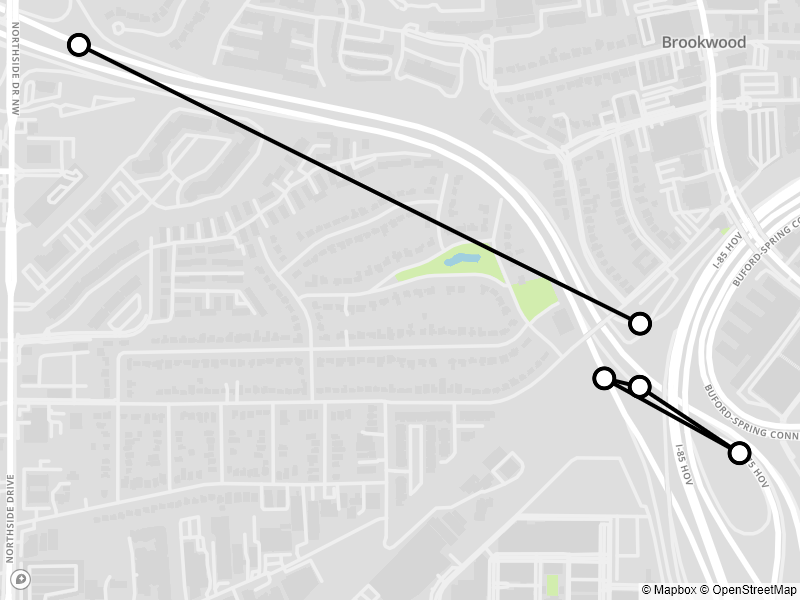}
    \caption{Traffic graph}
    \end{subfigure}
    \hspace{0.02in}
    \begin{subfigure}[h]{.4\linewidth}
    \includegraphics[width=1\linewidth]{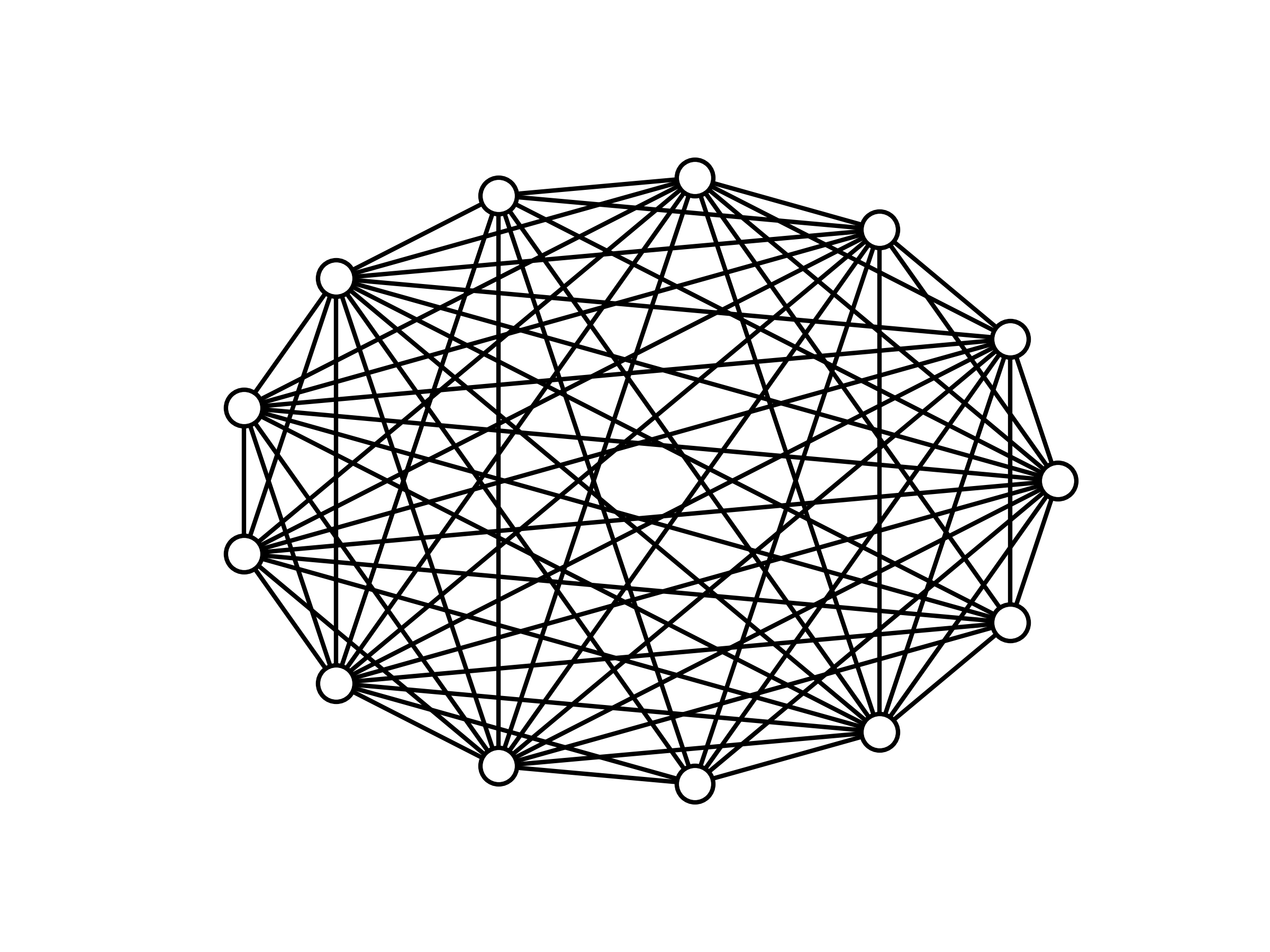}
    \caption{Sepsis graph}
    \end{subfigure}
    
\caption{Latent graph structures for the real data sets. From left to right, the graphs correspond to the ones in Atlanta traffic congestion data and the Sepsis data.}
\label{fig:real-data-graphs-app}
\end{figure}

\paragraph{Real data.} We also apply the model to four real data sets across different real-world settings. 

\begin{itemize}

    \item[(i)] {\it Traffic congestion data}. The Georgia Department of Transportation (GDOT) provides traffic volume data for sensors embedded on roads and highways in the state. We have access to such data for 5 sensors at the interchange of interstates 75 and 85 in Midtown Atlanta from September 2018 to March 2019. Traffic volume is measured in 15-minute intervals, and congestion events are detected when the traffic count exceeds the third quartile of the daily traffic volume. The result is 3,830 events which are split into 24-hour trajectories (with an average of 24 events per day). 

    \item[(ii)] {\it Sepsis data}. The Sepsis data set released by PhysioNet Challenges \citep{goldberger2000physiobank, reyna2019early} is a physiological data set that records patient covariates consisting of demographics, vital signs, and laboratory values for ICU patients from three separate hospital systems. Based on the covariates, common and clinically relevant Sepsis-Associated Derangements (SADs) can be identified through expert opinions. A SAD is considered present if the patient’s covariates are outside of normal limits \citep{wei2023granger}. In our Sepsis data, we extract the onsets of 12 SADs and the Sepsis (total of 13 medical indices, as shown in Table~\ref{tab:sepsis-medical-indices}) as discrete events, with a total number of 80,463 events. Each sequence contains the events extracted from one patient during a 24-hour period. The event times are measured in hours, with the average length of each sequence being 15.6.

    \item[(iii)] {\it Wildfire data}. The California Public Utilities Commission (CPUC) maintains a large-scale multi-modal wildfire incident dataset. We extract a total of 2,428 wildfire occurrences in California from 2014 to 2019. The latitude-longitude coordinates of incidents are bounded by the rectangular region [34.51, -123.50] $\times$ [40.73, -118.49]. Note that the majority of the region has no fire in the 5-year horizon due to the fact that fire incidents are likely to cluster in space. Therefore, we apply the K-means algorithm to extract 25 clusters of wildfire incidents. The entire dataset is split into one-year sequences with an average length of 436 events.

    \item[(iv)] {\it Theft data}. The proprietary crime data collected by the police department in Valencia, Spain, records the crime incidents that happened from 2015 to 2019, including incident location, crime category, and distance to various landmarks within the city. We analyze 9,372 sustraccions (smooth thefts) that happened near 52 pubs within the Valencia city area. Each sustraccion is assigned to the closest pub. We partition the data into quarter-year-long sequences with an average length of 469 events.

\end{itemize}

The learned latent graph structures overlaid on the real-world geography are displayed in Figure~\ref{fig:real-data-graphs}(a) and (b) for the California wildfire and Valencia theft data, respectively. The learned structure for the Atlanta traffic congestion and the Sepsis data is in Figure~\ref{fig:real-data-graphs-app}(a) and (b), respectively.

\begin{table}[!t]
  \caption{Identified 13 medical indices in Sepsis data.}
  \vspace{-0.1in}
  \centering
  \resizebox{1.\linewidth}{!}{
  \begin{threeparttable}
    \begin{tabular}{cccccccccccccc}
    \toprule
    \toprule
    Number & 1 & 2 & 3 & 4 & 5 & 6 & 7 & 8 & 9 & 10 & 11 & 12 & 13 \\
    \midrule
    Abbreviation & RI & EI & OCD & Shock & DCO & CI & Chl & HI & OD(Lab) & Inf(lab) & OD(vs) & Inf(vs) & Sepsis\\
    \multirow{2}{*}{Full name} & \multirow{2}{*}{\shortstack{Renal \\ Injury}} & \multirow{2}{*}{\shortstack{Electrolyte \\ Imbalance}} & \multirow{2}{*}{\shortstack{Oxygen Carrying \\ Dysfunction}} & \multirow{2}{*}{Shock} & \multirow{2}{*}{\shortstack{Diminished \\ Cardiac Output}} & \multirow{2}{*}{Coagulopathy} & \multirow{2}{*}{Cholestasis} & \multirow{2}{*}{\shortstack{Hepatocellular \\ Injury}} & \multirow{2}{*}{\shortstack{Oxygenation \\ Dysfunction (Lab)}} & \multirow{2}{*}{Inflammation (Lab)} & \multirow{2}{*}{\shortstack{Oxygenation Dysfunction \\ (vital sign)}} & \multirow{2}{*}{\shortstack{Inflammation \\ (vital sign)}} & \multirow{2}{*}{Sepsis} \\
    &  &  &  &  &  &  \\
    \bottomrule
    \bottomrule
  \end{tabular}
  \end{threeparttable}
  }
  \label{tab:sepsis-medical-indices}
\end{table}

\begin{table}[!t]
  \caption{Training hyper-parameters for the baselines.}
  \vspace{-0.1in}
  \centering
  \resizebox{1.\linewidth}{!}{
  \begin{threeparttable}
  \begin{tabular}{ccccccccccccccccccc}
    \toprule
    \toprule
    & \multicolumn{2}{c}{\multirow{2}{*}{\shortstack{ 3-node\\ synthetic}}}
    & \multicolumn{2}{c}{\multirow{2}{*}{\shortstack{ 16-node\\ synthetic}}}
    & \multicolumn{2}{c}{\multirow{2}{*}{\shortstack{ 50-node\\ synthetic}}}
    & \multicolumn{2}{c}{\multirow{2}{*}{\shortstack{ 50-node synthetic \\ with non-separable kernel}}}
    & \multicolumn{2}{c}{\multirow{2}{*}{\shortstack{ 225-node\\ synthetic}}}
    & \multicolumn{2}{c}{\multirow{2}{*}{\shortstack{ Traffic}}
    }
    & \multicolumn{2}{c}{\multirow{2}{*}{\shortstack{ Sepsis}}}
    & \multicolumn{2}{c}{\multirow{2}{*}{\shortstack{ Wildfire}}}
    & \multicolumn{2}{c}{\multirow{2}{*}{\shortstack{ Theft}}} \\
    &  &  &  &  &  &  &  & \\
    
    \cmidrule(lr){2-3} \cmidrule(lr){4-5}  \cmidrule(lr){6-7} \cmidrule(lr){8-9} \cmidrule(lr){10-11} \cmidrule(lr){12-13} \cmidrule(lr){14-15} \cmidrule(lr){16-17} \cmidrule(lr){18-19}
    
    \multirow{2}{*}{Model} & \multirow{ 2}{*}{\shortstack{Learning\\Rate}} & \multirow{ 2}{*}{\shortstack{Batch\\Size}} & \multirow{ 2}{*}{\shortstack{Learning\\Rate}} & \multirow{ 2}{*}{\shortstack{Batch\\Size}} & \multirow{ 2}{*}{\shortstack{Learning\\Rate}} & \multirow{ 2}{*}{\shortstack{Batch\\Size}} & \multirow{ 2}{*}{\shortstack{Learning\\Rate}} & \multirow{ 2}{*}{\shortstack{Batch\\Size}} & \multirow{ 2}{*}{\shortstack{Learning\\Rate}} & \multirow{ 2}{*}{\shortstack{Batch\\Size}} & \multirow{ 2}{*}{\shortstack{Learning\\Rate}} & \multirow{ 2}{*}{\shortstack{Batch\\Size}} & \multirow{ 2}{*}{\shortstack{Learning\\Rate}} & \multirow{ 2}{*}{\shortstack{Batch\\Size}} & \multirow{ 2}{*}{\shortstack{Learning\\Rate}} & \multirow{ 2}{*}{\shortstack{Batch\\Size}} & \multirow{ 2}{*}{\shortstack{Learning\\Rate}} & \multirow{ 2}{*}{\shortstack{Batch\\Size}} \\
    &  &  &  &  &  &  \\
    \midrule
    \texttt{RMTPP}       & $10^{-3}$  & $32$ & $10^{-3}$  &  $32$ & $10^{-3}$ & $32$ & $10^{-3}$ & $32$ & $10^{-3}$ & $32$ & $10^{-3}$ & $32$ & $10^{-3}$ & $100$ & $10^{-3}$  & $2$ & $10^{-3}$ & $2$ \\
    \texttt{FullyNN}     & $10^{-2}$ & $100$ & $10^{-2}$ & $100$ & $10^{-2}$ & $100$ & $10^{-2}$ & $100$ & $10^{-2}$ & $100$ & $10^{-3}$ & $50$ & $10^{-3}$ & $100$ & $10^{-3}$ & $20$  & $10^{-3}$ & $100$ \\
    \texttt{DNSK}        & $10^{-1}$ & $32$  & $10^{-1}$ & $32$ & $10^{-1}$ & $32$ & $10^{-1}$ & $32$ & $10^{-1}$ & $32$ & $10^{-1}$ & $32$ & $10^{-1}$ & $100$ & $10^{-1}$ & $2$ & $10^{-1}$ & $2$ \\
    \texttt{THP-S}       & $10^{-3}$ & $32$  & $10^{-3}$  & $32$  & $10^{-3}$ & $32$ & $10^{-3}$ & $32$ & $10^{-3}$ & $32$ & $10^{-3}$ & $64$  & $10^{-3}$ & $100$ & $10^{-3}$ & $2$ & $10^{-3}$ & $2$ \\
    \texttt{SAHP-G}      & $10^{-4}$  & $32$  & $10^{-4}$  & $32$  & $10^{-4}$ & $32$ & $10^{-4}$ & $32$ & $10^{-4}$ & $32$ & $10^{-4}$ & $32$ & $10^{-3}$ & $100$ & $10^{-3}$ & $2$ & $10^{-3}$ & $2$ \\
    \bottomrule
    \bottomrule
  \end{tabular}
  \end{threeparttable}
  }
  \vspace{-0.05in}
  \label{tab:hyper-parameters}
\end{table}

\subsection{Detailed experimental setup}

We choose our temporal basis functions to be fully connected neural networks with two hidden layers of width 32. Each layer, except for the output layer, is equipped with the softplus activation function. 
The \texttt{GraDK} method sets the kernel rank to be $L=1$ and $R=3$ for the final learned model for each data set (except $L=1, R=2$ on the data generated by the 50-node graph, $L=2, R=4$ on the data generated by the 50-node graph with the non-separable kernel, and $L=2, R=4$ on the Sepsis data).
For each data set, all the models are trained using 80\% of the data and tested on the remaining 20\%. Our model parameters are estimated through objective functions in Section~\ref{sec:model-estimation} using the Adam optimizer~\citep{kingma2014adam} with a learning rate of $10^{-2}$ and batch size of $32$.

For the baseline of \texttt{RMTPP}, we test the dimension of \{32, 64, 128\} for the hidden embedding in RNN and choose an embedding dimension of 32 in the experiments. For \texttt{FullyNN}, we set the embedding dimension to be 64 and use a fully-connected neural network with two hidden layers of width 64 for the cumulative hazard function, as the default ones in the original paper. The dimension of input embedding is set to 10. For \texttt{DNSK}, we adopt the structure for the marked point processes (for spatio-temporal point processes in the ablation study) and set the rank of temporal basis and mark basis (spatial basis in the ablation study) to one and three. \texttt{THP-S} and \texttt{SAHP-G} use the default parameters from their respective code implementations. Learning rate and batch size parameters are provided for each baseline and experiment in Table~\ref{tab:hyper-parameters}.

We note that the original paper of \texttt{FullyNN} \citep{omi2019fully} does not consider the modeling of event marks, and the model can only be used for temporal point processes, for it requires the definition of cumulative intensity and the calculation of the derivative, which is inapplicable in event mark space. 
To compare with this method in the setting of graph point processes, we extend \texttt{FullyNN} to model the mark distribution. Specifically, an additional feedforward fully-connected neural network $\Psi$ is introduced. For each index $i$, it takes the hidden states of history up to $i$-th event $\bm{h}_i$ as input and outputs its prediction for the likelihood of the next mark $v_{i+1}$ as $\Psi(v_{i+1}|\bm{h}_i)$. Given a sequence of $n$ events $\{(t_i, v_i)\}_{i=1}^{n}$, the optimization objective of model log-likelihood for the extended \texttt{FullyNN} is written as:
\[
\begin{aligned}
    \mathcal{L} &= \text{log}L(\{t_i\}) + \text{log}L(\{v_i\}) \\
    &= \sum_i \left[\text{log}\left\{\frac{\partial}{\partial\tau}\Phi(\tau=t_{i+1} - t_i|\bm{h}_i)\right\} - \Phi(t_{i+1}-t_i|\bm{h}_i) \right] + \sum_i\text{log}\left\{\Psi(v_{i+1}|\bm{h}_i)\right\},
\end{aligned}
\]
where $\Phi$ is the cumulative hazard function (\textit{i.e.}, cumulative intensity function) introduced in \texttt{FullyNN}, parametrized by a feedforward neural network.
Such an approach to modeling the distribution of event marks has been commonly adopted in previous studies \citep{du2016recurrent, shchur2019intensity}.

\begin{table*}[!t]
  \caption{Additional synthetic data results.}
  \vspace{-0.1in}
  \centering
  \resizebox{1.\linewidth}{!}{
  \begin{threeparttable}
    \begin{tabular}{cccccccccc}
    \toprule
    \toprule
    & \multicolumn{3}{c}{\multirow{2}{*}{\shortstack{3-node graph\\with \textit{negative} influence}}} & \multicolumn{3}{c}{\multirow{2}{*}{50-node graph}} & \multicolumn{3}{c}{\multirow{2}{*}{225-node graph}} \\ \\
    \cmidrule(lr){2-4} \cmidrule(lr){5-7}  \cmidrule(lr){8-10}
    \multirow{2}{*}{Model} & \multirow{2}{*}{\shortstack{Testing \\ $\ell$}} & \multirow{2}{*}{\shortstack{Time \\ MAE}} & \multirow{2}{*}{\shortstack{Type \\ KLD}} & \multirow{2}{*}{\shortstack{Testing \\ $\ell$}} & \multirow{2}{*}{\shortstack{Time \\ MAE}} & \multirow{2}{*}{\shortstack{Type \\ KLD}} & \multirow{2}{*}{\shortstack{Testing \\ $\ell$}} & \multirow{2}{*}{\shortstack{Time \\ MAE}} & \multirow{2}{*}{\shortstack{Type \\ KLD}} \\ \\
    \midrule
    \texttt{RMTPP} & $-3.473_{(0.087)}$ & $0.528$ & $0.093$ & $-27.915_{(1.251)}$ & $37.666$ & $0.103$ & $-16.294_{(1.047)}$ & $29.329$ & $0.252$ \\
    \texttt{FullyNN}  & $-2.086_{(0.009)}$ & $0.291$ & $0.006$ & $-1.736_{(0.019)}$ & $13.295$ & $0.058$ & $-3.241_{(0.026)}$ & $9.672$ & $0.174$ \\
    \texttt{DNSK-mtpp} & $-2.127_{(0.003)}$ & $0.149$ & $0.012$ &$-1.165_{(0.003)}$ & $1.074$ & $0.076$ & $-2.511_{(0.001)}$ & $3.958$ & $0.086$\\
    \midrule
    \texttt{THP-S}  & $-2.089_{(0.008)}$ & $0.413$ & $0.006$ & $-1.091_{(0.005)}$ & $3.940$ & $0.019$ & $-2.550_{(0.008)}$ & $4.109$ & $0.087$ \\
    \texttt{SAHP-G}  & $-2.113_{(0.005)}$ & $0.172$ & $0.003$ &$-1.099_{(0.004)}$ & $1.119$ & $0.014$ & $-2.506_{(0.005)}$ & $3.578$ & $0.094$\\
    \midrule
    \texttt{GraDK+L3net+NLL} & $-2.062_{(0.003)}$ & $0.048$ & $\textbf{<0.001}$ & $-1.065_{(0.005)}$ & $1.023$ & $\textbf{0.004}$ & $-2.487_{(0.005)}$ & $1.779$ & $0.015$ \\
    \texttt{GraDK+L3net+LS} & $\textbf{--2.059}_{(0.002)}$ & $\textbf{0.021}$ & $\textbf{<0.001}$ & $\textbf{--1.056}_{(0.003)}$ & $\textbf{0.957}$ & $0.005$ & $\textbf{--2.485}_{(0.003)}$ & $\textbf{0.146}$ & $\textbf{0.012}$\\
    \texttt{GraDK+GAT+NLL} & $-2.073_{(0.004)}$ & $0.068$ & $\textbf{<0.001}$ & $-1.690_{(0.001)}$ & $0.690$ & $0.007$ & $-2.493_{(0.006)}$ & $2.458$ & $0.092$ \\
    \bottomrule
    \bottomrule
  \end{tabular}
  \begin{tablenotes}
  \item *Numbers in parentheses are standard errors for three independent runs.
  \end{tablenotes}
  \end{threeparttable}
  }
  \label{tab:app-synthetic-data-results}
\end{table*}

\subsection{Additional experimental results}

\begin{figure}[!t]
    \centering
    \begin{subfigure}[h]{.99\linewidth}
    {\includegraphics[width=\linewidth]{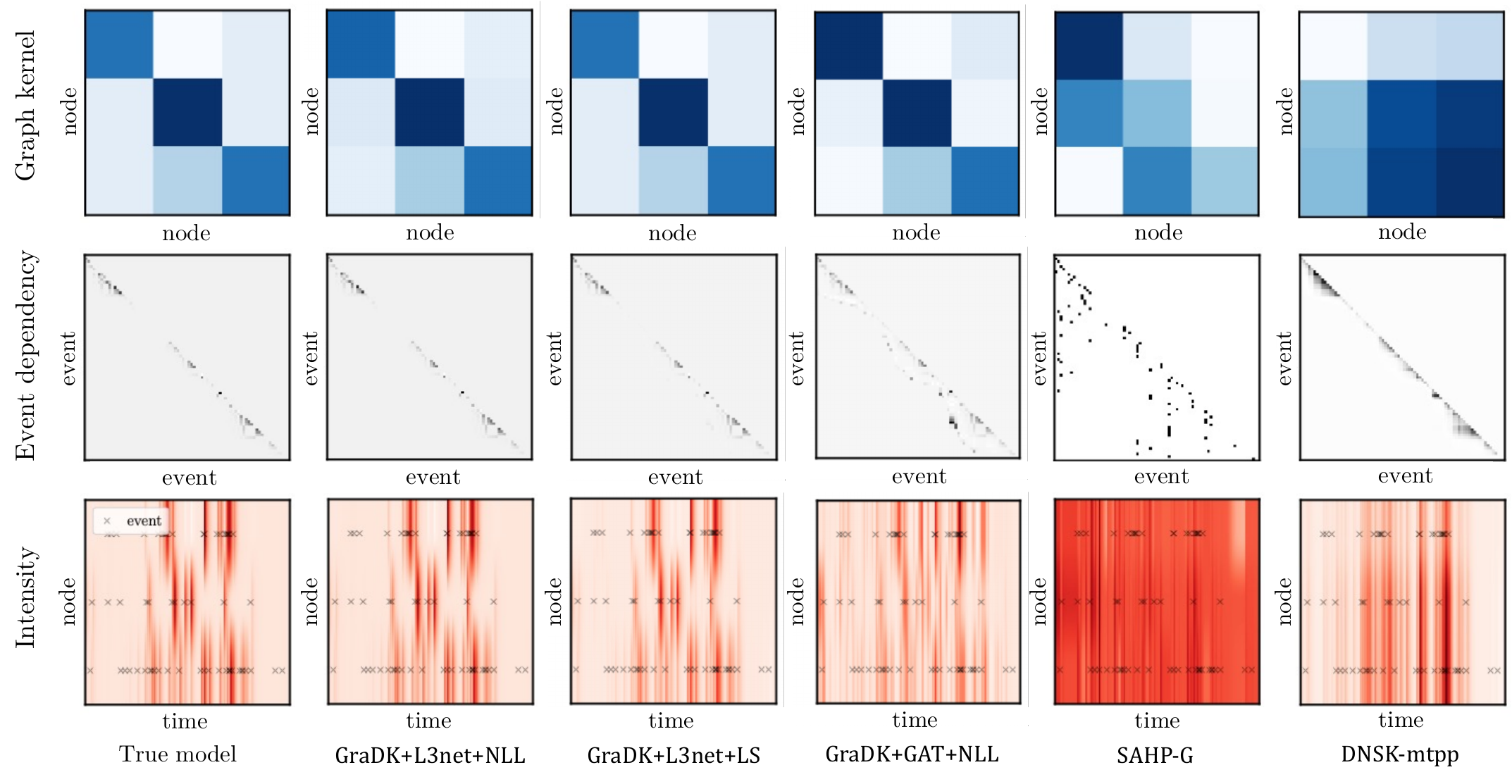}}
    \end{subfigure}
\caption{Graph kernel, inter-event dependence, and conditional intensity recovery for the synthetic data set generated by the 3-node graph with negative (inhibitive) graph influence. The first column reflects the ground truth, while the subsequent columns reflect the results obtained by \texttt{GraDK} (our method), \texttt{SAHP-G}, and \texttt{DNSK}, respectively.}
\vspace{-0.1in}
\label{fig:3-node-synthetic-data-vis}
\end{figure}

\begin{figure}[!t]
    \centering
    \begin{subfigure}[h]{.99\linewidth}
    {\includegraphics[width=\linewidth]{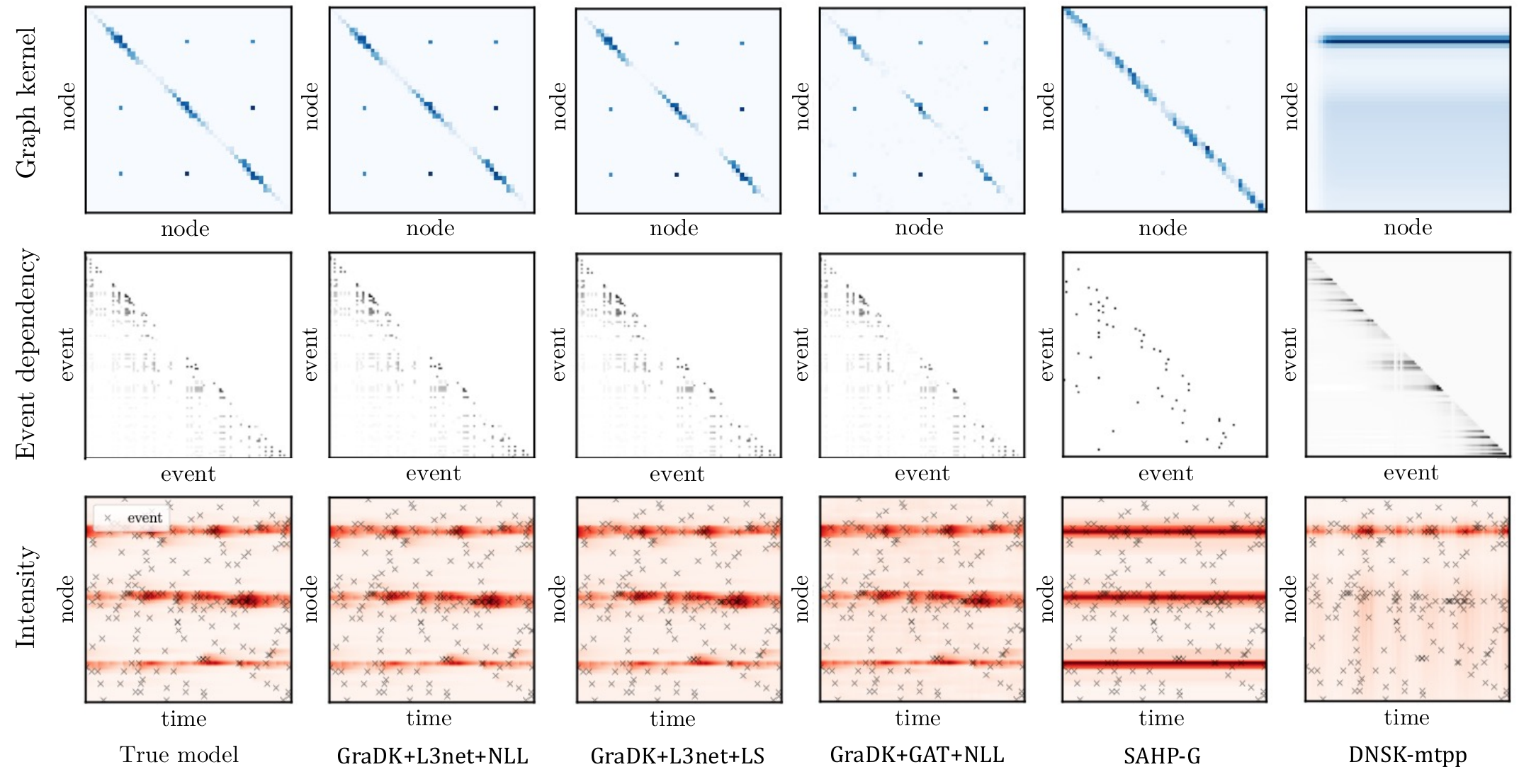}}
    \end{subfigure}
\caption{Graph kernel, inter-event dependence, and conditional intensity recovery for the synthetic data set generated by the 50-node graph.}
\label{fig:50-node-synthetic-data-vis}
\end{figure}

\begin{figure}[!t]
    \centering
    \begin{subfigure}[h]{.99\linewidth}
    {\includegraphics[width=\linewidth]{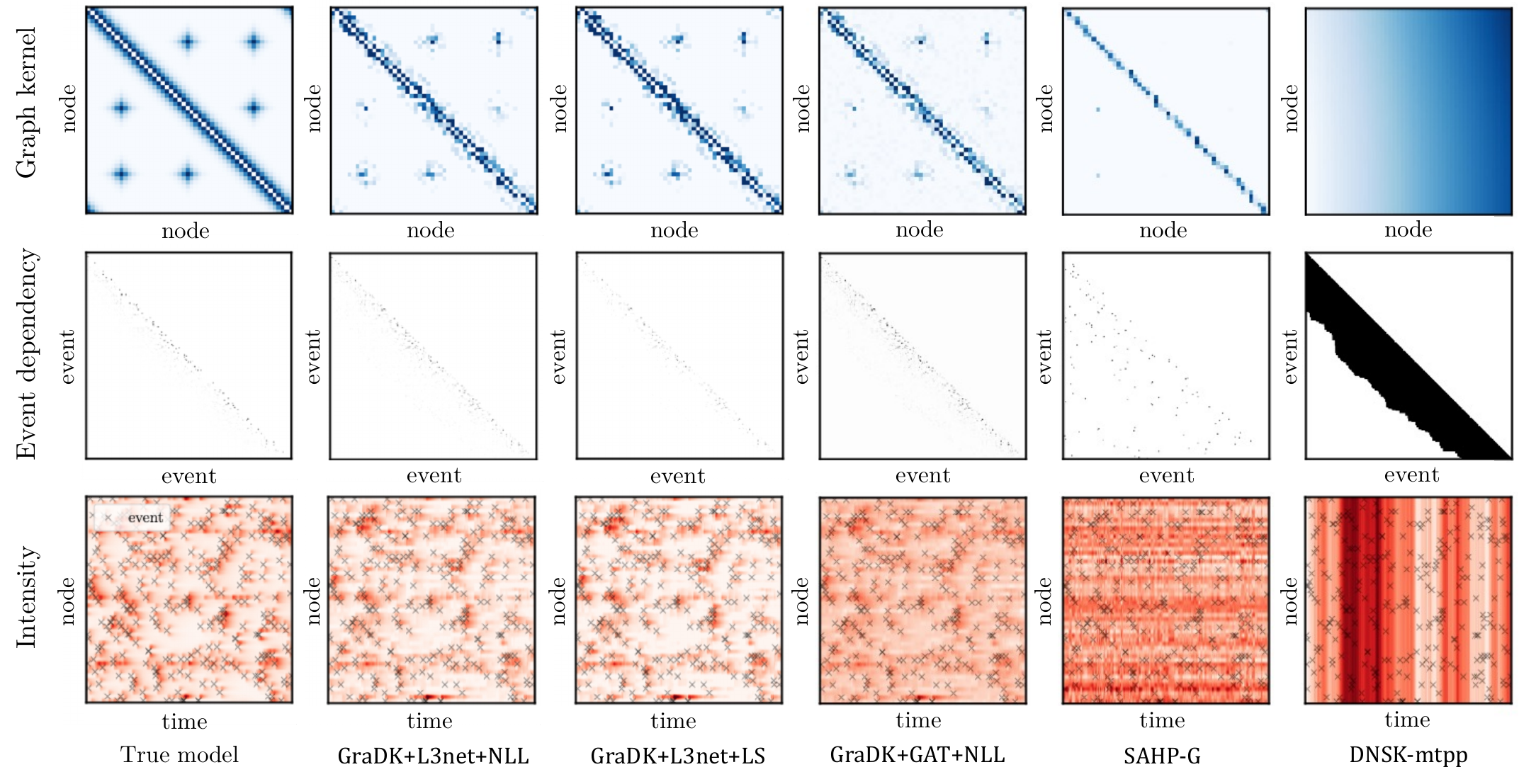}}
    \end{subfigure}
\caption{Graph kernel, inter-event dependence, and conditional intensity recovery for the synthetic data set generated on the 50-node graph with a non-separable kernel.}
\label{fig:non-separable-kernel-synthetic-data-vis}
\end{figure}

\begin{figure}[!t]
    \centering
    \begin{subfigure}[h]{.99\linewidth}
    {\includegraphics[width=\linewidth]{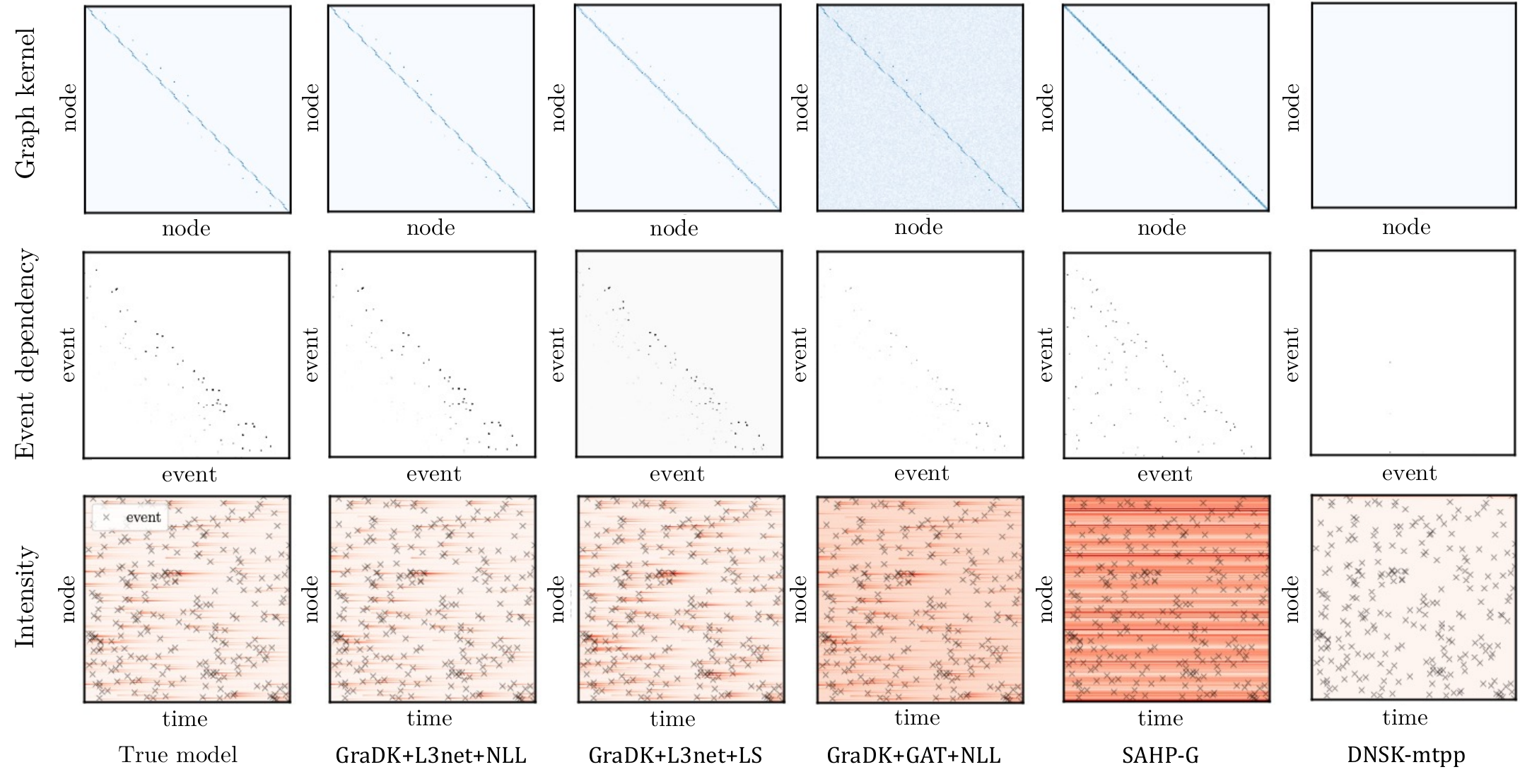}}
    \end{subfigure}
\caption{Graph kernel, inter-event dependence, and conditional intensity recovery for the 225-node synthetic data set.}
\label{fig:225-node-synthetic-data-vis}
\end{figure}

\paragraph{Synthetic data.} \rev{See Table~\ref{tab:app-synthetic-data-results} for quantitative results in synthetic datasets 1, 3, and 5.} Figure~\ref{fig:3-node-synthetic-data-vis} presents the kernel and intensity recovery results for the synthetic data set 1 generated by a non-stationary temporal kernel and a 3-node-graph kernel with inhibitive influence. The recovery outcomes demonstrate the efficacy of our proposed model in characterizing the temporal and graph-based dependencies among events, as evident from the first and second rows of the figure. Moreover, the results emphasize the capability of our model, \texttt{GraDK}, to effectively capture the inhibitive effects of past events. In contrast, the event dependencies represented by the normalized positive attention weights in \texttt{SAHP-G} solely capture the triggering intensity of past events without accounting for inhibitive influences. Similarly, the first row of Figure~\ref{fig:50-node-synthetic-data-vis} displays the true graph influence kernel in the 50-node synthetic data set and the learned graph kernels by \texttt{GraDK}, \texttt{SAHP-G}, and \texttt{DNSK}. While \texttt{SAHP-G} exaggerates the self-exciting influence of graph nodes and \texttt{DNSK} only learns some semblance of the graph kernel behavior, our method accurately recovers both self- and inter-node influence patterns, resulting in a faithful model of the true graph kernel. 
The conditional intensity via each method for one testing trajectory is displayed in the third row of Figure~\ref{fig:50-node-synthetic-data-vis}, which demonstrates the capability of our model to capture the temporal dynamics of events.
The same results are presented in Figure~\ref{fig:non-separable-kernel-synthetic-data-vis} and Figure~\ref{fig:225-node-synthetic-data-vis} for the synthetic data sets generated by the 50-node graph with non-separable kernel and 225-node graph, respectively.

\begin{figure}[!t]
    \begin{subfigure}[h]{1.\linewidth}
    \centering
    \includegraphics[width=.24\linewidth]{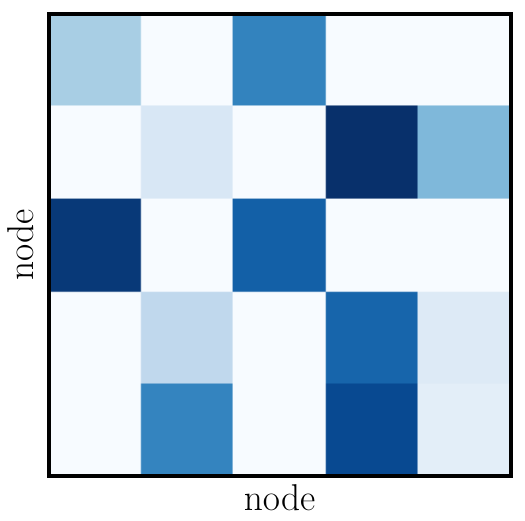}
    \includegraphics[width=.24\linewidth]{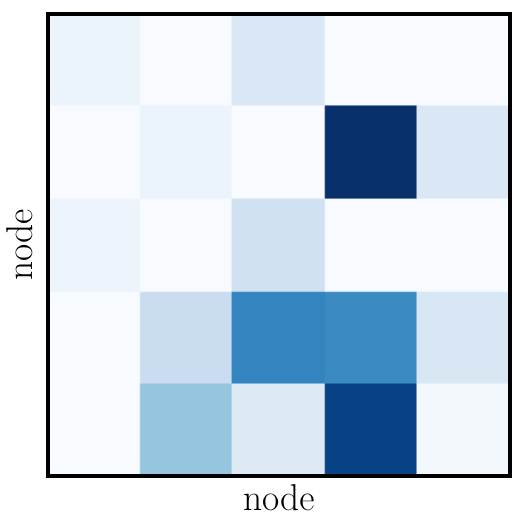}
    \includegraphics[width=.24\linewidth]{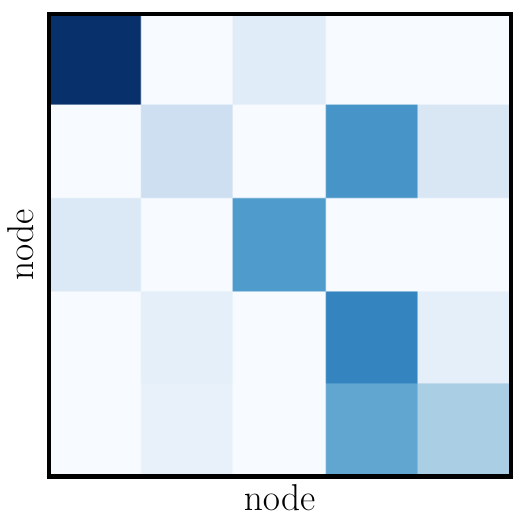}
    \includegraphics[width=.24\linewidth]{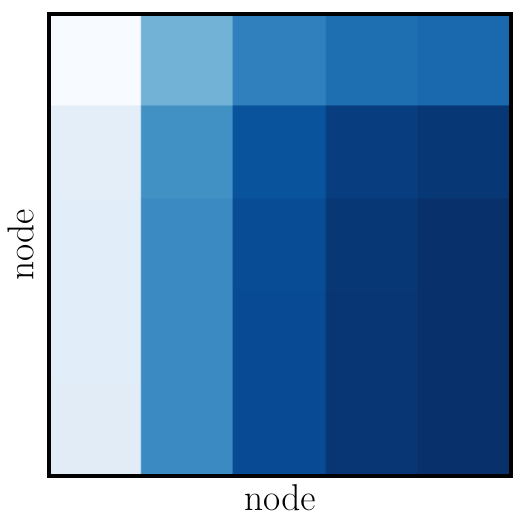}
    \caption{Traffic data}
    \end{subfigure}

    \begin{subfigure}[h]{1.\linewidth}
    \centering
    \includegraphics[width=.24\linewidth]{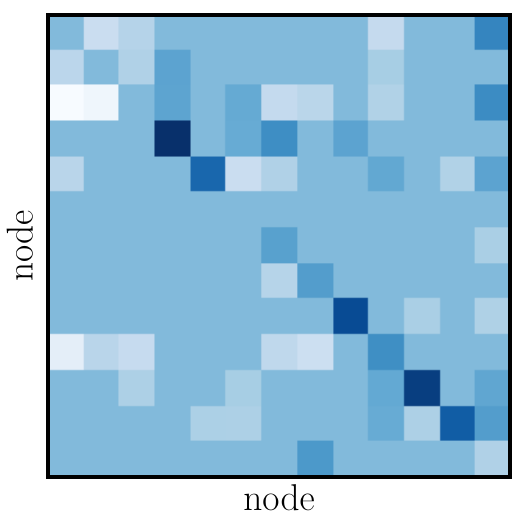}
    \includegraphics[width=.24\linewidth]{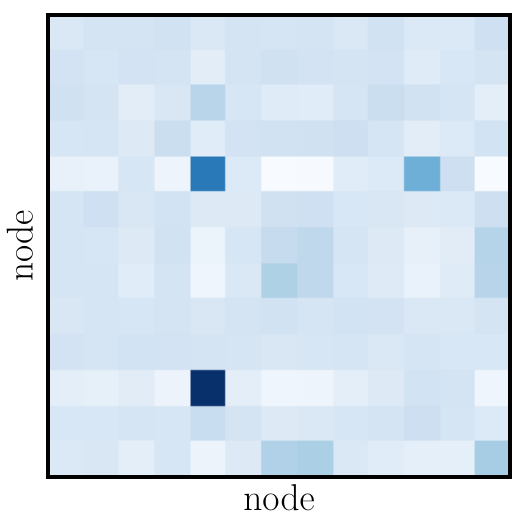}
    \includegraphics[width=.24\linewidth]{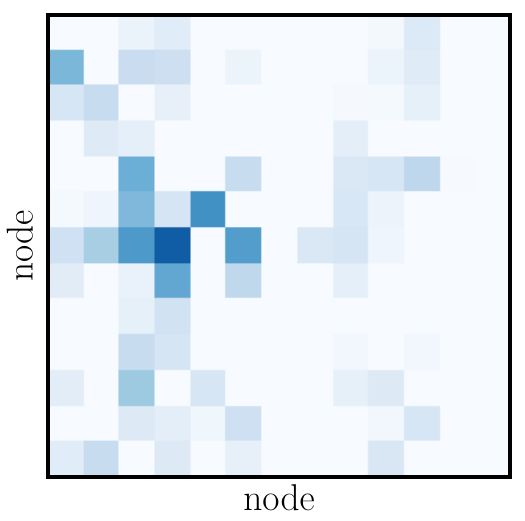}
    \includegraphics[width=.24\linewidth]{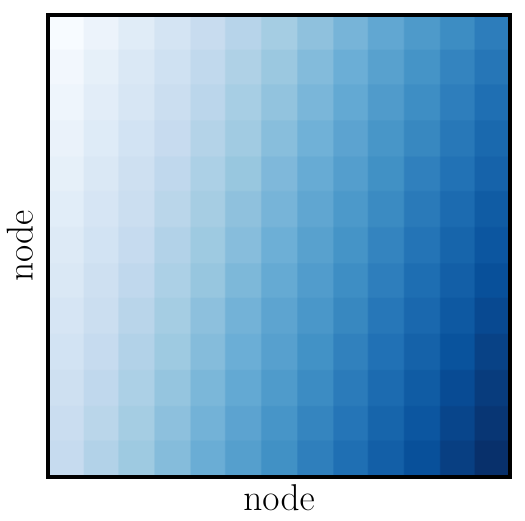}
    \caption{Sepsis data}
    \end{subfigure}

    \begin{subfigure}[h]{1.\linewidth}
    \centering
    \includegraphics[width=.24\linewidth]{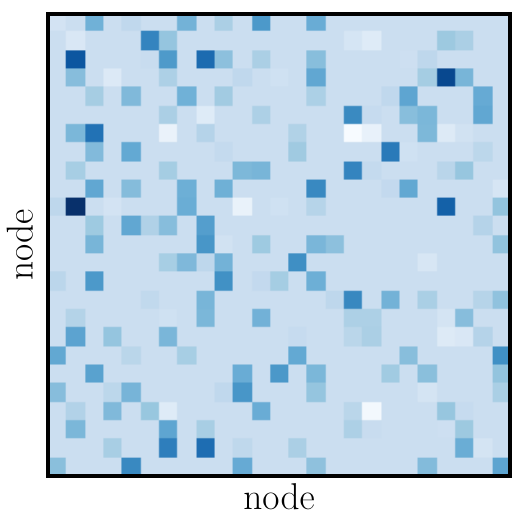}
    \includegraphics[width=.24\linewidth]{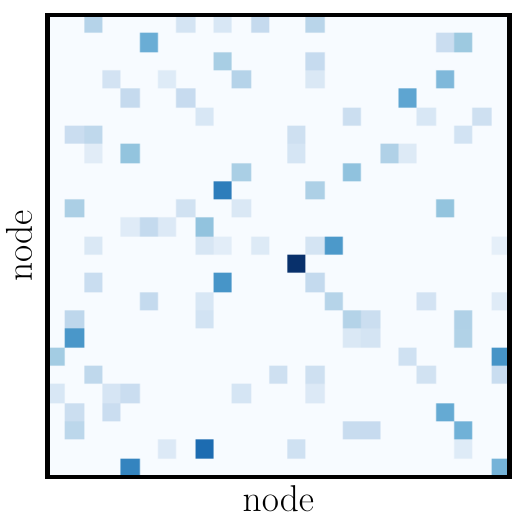}
    \includegraphics[width=.24\linewidth]{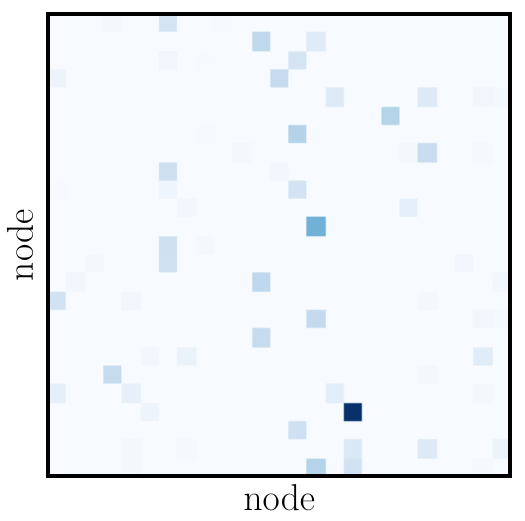}
    \includegraphics[width=.24\linewidth]{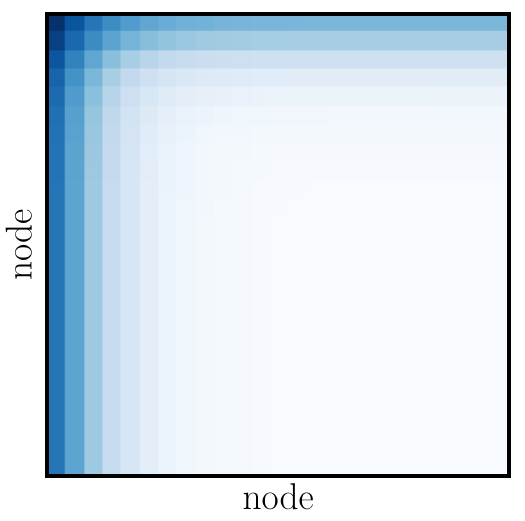}
    \caption{Wildfire data}
    \end{subfigure}
\caption{Learned graph kernels for traffic, Sepsis, and wildfire data set evaluated at $t'=0$ and $t-t'=1$. The columns present the recovered kernels on each data set of \texttt{GraDK+L3net}, \texttt{GraDK+GAT}, \texttt{SAHP-G}, and \texttt{DNSK}, respectively.}
\label{fig:add-real-data-vis}
\end{figure}

\begin{table*}[!t]
  \caption{Additional real data results.}
  \vspace{-0.1in}
  \centering
  \begin{threeparttable}
  \begin{tabular}{ccccccc}
    \toprule
    \toprule
    & \multicolumn{3}{c}{\multirow{2}{*}{\shortstack{Traffic congestion\\(5 nodes)}}} & \multicolumn{3}{c}{\multirow{2}{*}{\shortstack{Wildfire\\(25 nodes)}}} \\ \\
    \cmidrule(lr){2-4} \cmidrule(lr){5-7}
    \multirow{2}{*}{Model} & \multirow{2}{*}{\shortstack{Testing \\ $\ell$}} & \multirow{2}{*}{\shortstack{Time \\ MAE}} & \multirow{2}{*}{\shortstack{Type \\ KLD}} & \multirow{2}{*}{\shortstack{Testing \\ $\ell$}} & \multirow{2}{*}{\shortstack{Time \\ MAE}} & \multirow{2}{*}{\shortstack{Type \\ KLD}} \\ \\
    \midrule
    \texttt{RMTPP} & $-5.197_{(0.662)}$ & $2.348$ & $0.021$ & $-6.155_{(1.589)}$ & $1.180$ & $0.178$ \\
    \texttt{FullyNN} & $-3.292_{(0.108)}$ & $0.511$ & $0.012$ & $-4.717_{(0.119)}$ & $0.817$ & $0.026$ \\
    \texttt{DNSK-mtpp} & $-2.401_{(0.011)}$ & $0.934$ & $0.010$ & $-3.706_{(0.008)}$ & $0.711$ & $0.083$ \\
    \midrule
    \texttt{THP-S} & $-2.254_{(0.007)}$ & $0.378$ & $0.003$ & $-4.523_{(0.018)}$ & $1.183$ & $0.134$ \\
    \texttt{SAHP-G} & $-2.453_{(0.013)}$ & $0.729$ & $0.021$ & $-3.919_{(0.040)}$ & $0.551$ & $0.032$ \\
    \midrule
    \texttt{GraDK+L3net+NLL} & $-2.178_{(0.005)}$ & $0.314$ & $\textbf{0.001}$ & $--3.625_{(0.002)}$ & $0.207$  & $0.006$ \\
    \texttt{GraDK+L3net+LS} & $--2.159_{(0.004)}$ & $0.247$ & $0.001$ & $-3.628_{(0.002)}$ & $0.347$ & $0.013$ \\
    \texttt{GraDK+GAT+NLL} & $-2.281_{(0.011)}$ & $0.356$ & $0.015$ & $-3.629_{(0.007)}$ & $0.898$ & $0.085$ \\
    \bottomrule
    \bottomrule
  \end{tabular}
  \begin{tablenotes}
  \item *Numbers in parentheses are standard errors for three independent runs.
  \end{tablenotes}
  \end{threeparttable}
  \label{tab:app-real-data-results}
\end{table*}

\paragraph{Real data.} \rev{See Table~\ref{tab:app-real-data-results} for quantitative results in the traffic and wildfire data.} We visualize the learned graph kernels by \texttt{GraDK}, \texttt{SAHP-G}, and \texttt{DNSK} on traffic, Sepsis, and wildfire data in Figure~\ref{fig:add-real-data-vis}. Our approach is able to learn intense graph signals with complex patterns by taking advantage of the graph structure and GNNs. While the attention structures adopted in \texttt{SAHP-G} contribute to improved model prediction performance for future events, this approach suffers from the limited model expressiveness and interpretability when attempting to recover the underlying mechanism of the event generation process, indicated by the weak and noisy graph signals. \texttt{DNSK} fails to uncover the intricate patterns existing in graph kernels and only provides restrictive kernels for event modeling without considering the latent graph structures among data.

\begin{figure}[!t]
    \centering
    \begin{subfigure}[h]{.315\linewidth}
    \includegraphics[width=1\linewidth]{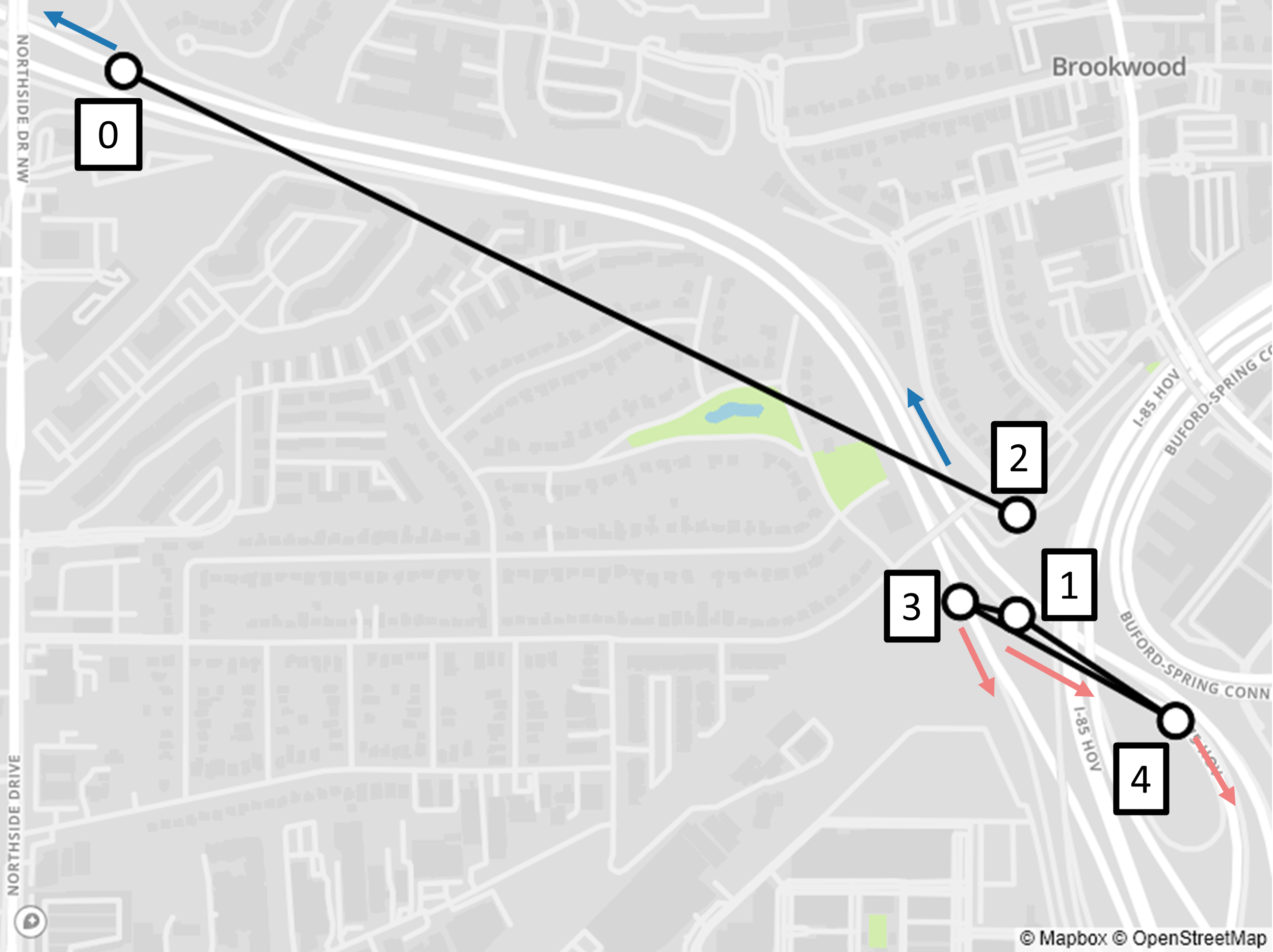}
    \caption{Traffic network structure}
    \end{subfigure}
    \hspace{0.1in}
    \begin{subfigure}[h]{.3\linewidth}
    \includegraphics[width=1\linewidth]{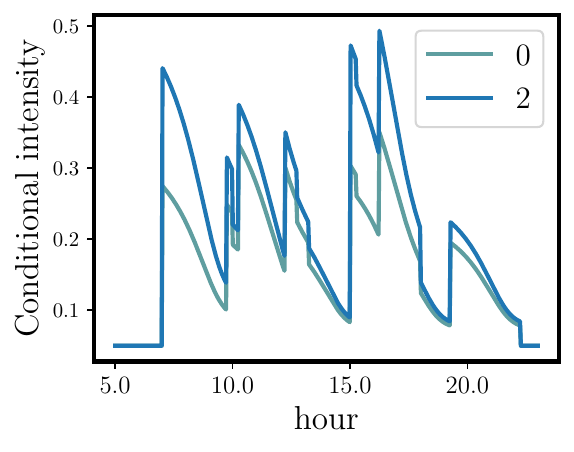}
    \caption{Northbound sites}
    \end{subfigure}
    \begin{subfigure}[h]{.3\linewidth}
    \includegraphics[width=1\linewidth]{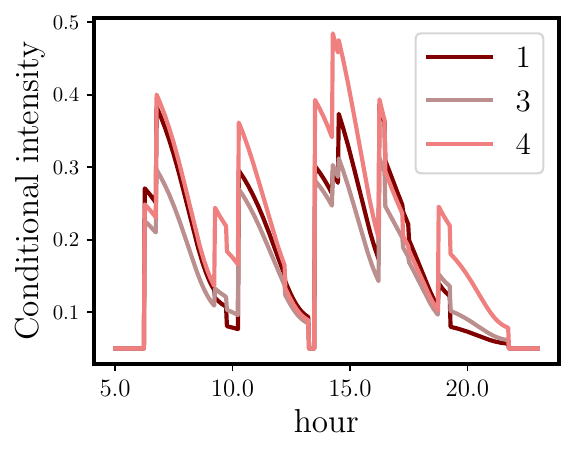}
    \caption{Southbound sites}
    \end{subfigure}
\caption{(a) Traffic network structure. Each traffic sensor is labeled with a number, and arrows indicate the monitored traffic directions of sensors on northbound and southbound highways, respectively. (b)(c) Conditional intensity of five sensors in a single day, grouped by monitored traffic direction.}
\label{fig:traffic-sensor-intensity}
\end{figure}

Our model not only achieves exceptional interpretability but also holds practical significance in the context of real-world point process data modeling. This can be demonstrated through experimental results conducted on traffic data. The five traffic sensors from which we collect data can be categorized into two groups, northbound sites, and southbound sites, according to the directions of the highway they are monitoring. Figure~\ref{fig:traffic-sensor-intensity}(a) visualizes the structure of the traffic network with sensors (graph nodes) labeled and arrows indicating the highway directions. We then investigate the conditional intensity learned by \texttt{GraDK} on each traffic sensor. Figure~\ref{fig:traffic-sensor-intensity}(b)(c) show the conditional intensity functions of five sensors during one single day (\textit{i.e.}, computed given one sequence from the testing set) with two subgroups of northbound and southbound sites. Note that similar temporal patterns can be found within the same subgroup, which can be attributed to the fact that the sensors in the same group are in the same direction and share the same traffic flow. Also, all the intensity functions show a temporal pattern in which they reach pinnacles during the morning (around 8:00) and evening (around 17:00) rush hours, with a higher possibility for traffic congestion at southbound sensors in the afternoon.

\subsection{Additional discussion of experimental results}
\label{app:exp-discussion}
\revv{This subsection complements the streamlined discussion in the main text.}

\paragraph{Likelihood versus prediction.}
\revv{In some real-data settings, attention-based models (e.g., \texttt{SAHP-G}) achieve slightly higher testing log-likelihood, while \texttt{GraDK} consistently yields substantially better prediction performance (Time MAE and Type KLD). This illustrates that likelihood maximization and predictive accuracy are not equivalent objectives for flexible neural point process models, and we view prediction accuracy as a primary practical objective in many point process applications.}

\paragraph{Interpretation of learned kernels.}
\revv{Figures~\ref{fig:real-data-vis} and~\ref{fig:add-real-data-vis} suggest that \texttt{GraDK} recovers structured multi-hop influence patterns, whereas \texttt{DNSK} is less informative in discrete graph settings and \texttt{SAHP-G} yields weaker kernel structure. This is consistent with attention models fitting local temporal patterns more aggressively, while the proposed kernel-based approach emphasizes structured graph influence that improves generalization.}

\subsection{Benefits of kernel parametrization with displacement}

\begin{table*}[!t]
  \caption{Comparison of kernels parameterized with and without displacement.}
  \vspace{-0.1in}
  \centering
  \begin{threeparttable}
    \begin{tabular}{ccccccc}
    \toprule
    \toprule
    \multirow{2}{*}{} & \multicolumn{3}{c}{\multirow{2}{*}{\shortstack{16-node graph \\ with \textit{2-hop} influence}}} & \multicolumn{3}{c}{\multirow{2}{*}{\shortstack{Wildfire \\ (25 nodes)}}} \\ \\
    \cmidrule(lr){2-4} \cmidrule(lr){5-7}
    \multirow{2}{*}{Model} & \multirow{2}{*}{\shortstack{Testing \\ $\ell$}} & \multirow{2}{*}{\shortstack{Time \\ MAE}} & \multirow{2}{*}{\shortstack{Type \\ KLD}} & \multirow{2}{*}{\shortstack{Testing \\ $\ell$}} & \multirow{2}{*}{\shortstack{Time \\ MAE}} & \multirow{2}{*}{\shortstack{Type \\ KLD}} \\ \\
    \midrule
    $k$ & $\textbf{--2.995}_{(0.004)}$ & $\textbf{0.028}$ & $\textbf{0.002}$ & $\textbf{--3.625}_{(0.002)}$ & $\textbf{0.207}$  & $\textbf{0.006}$ \\
    $k^{\rm{nodisp}}$ & $-3.026_{(0.005)}$ & $0.107$ & $\textbf{0.002}$ & $-3.707_{(0.004)}$ & $0.678$  & $0.009$ \\
    \bottomrule
    \bottomrule
  \end{tabular}
  \end{threeparttable}
  \vspace{-0.05in}
  \label{tab:benefits-of-displacement}
\end{table*}

\rev{The following ablation study is performed to prove the benefits of adopting the temporal kernel with displacement in kernel modeling \eqref{eq:temporal-graph-decomposed-kernel}. We replace the $g_d(t', t-t')$ in the proposed kernel with $g_d(t', t)$, leading to the following kernel without displacement:
\[
    \begin{aligned}
        k^{\rm{nodisp}}(t', t, v', v) &= \sum_{d=1}^D \sigma_d g_d(t', t) h_d(v', v) \\ 
        &= \sum_{r=1}^{R}\sum_{l=1}^{L}\alpha_{rl}\psi_l(t^\prime)\varphi_l(t)B_r(v^\prime, v),
    \end{aligned}
\]
We then fit this kernel on the synthetic data generated from the point process on the 16-node graph and the wildfire data, and compare its fitting and predicting performance with our kernel parameterized with displacement $t - t'$. We use graph kernel with L3net and NLL loss to fit both models. Thus, the only difference between these two models is the parametrization of the temporal kernel.}

\begin{figure}[!t]
    \centering
    \begin{subfigure}[h]{.8\linewidth}
    {\includegraphics[width=\linewidth]{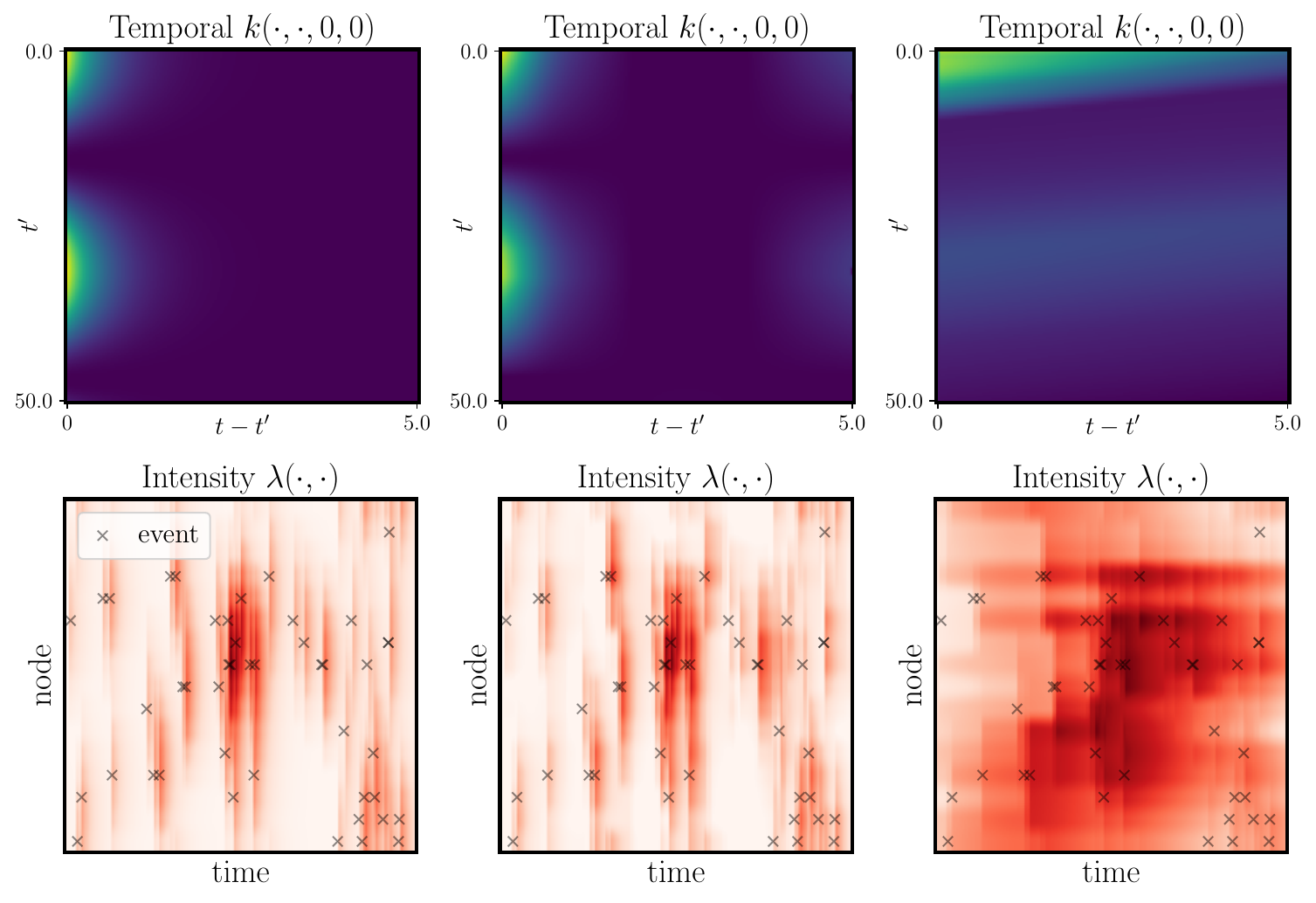}}
    \end{subfigure}
\caption{Visualization of the temporal kernel and the event intensity of the synthetic data on the 16-node graph. From left to right: true model, learned model by \texttt{GraDK+L3net+NLL}, and learned model by \texttt{GraDK+L3net+NLL} without displacement-based kernel parametrization.}
\label{fig:benefits-of-displacement-vis}
\end{figure}

\rev{
Table~\ref{tab:benefits-of-displacement} shows the results of data fitting and event prediction by the two kernels. Across both datasets, the model using the displacement-based kernel ($k$) consistently outperforms the variant without displacement ($k^{\rm{nodisp}}$) by showing a higher testing negative log-likelihood and smaller errors of the predicted times and types of future events. 
A visualization of the true model and the learned models in Figure~\ref{fig:benefits-of-displacement-vis} also indicates that the proposed \texttt{GraDK} can better recover the ground truth when equipped with the displacement-based kernel parametrization.
These results demonstrate the modeling benefits of incorporating displacement ($t - t'$) into the kernel, validating our kernel parametrization strategy in \eqref{eq:temporal-graph-decomposed-kernel}.
}

\end{document}